\documentclass[letterpaper]{article} 
\usepackage{aaai24}  
\usepackage{times}  
\usepackage{helvet}  
\usepackage{courier}  
\usepackage[hyphens]{url}  
\usepackage{graphicx} 
\usepackage{natbib}  
\usepackage{caption} 
\usepackage{algorithm}
\usepackage{algorithmic}
\usepackage{adjustbox}
\usepackage{newfloat}
\usepackage{listings}
\DeclareCaptionStyle{ruled}{labelfont=normalfont,labelsep=colon,strut=off} 
\floatstyle{ruled}
\newfloat{listing}{tb}{lst}{}
\floatname{listing}{Listing}
\usepackage{microtype}
\usepackage{graphicx}
\usepackage{caption}
\usepackage{subfigure}
\usepackage{booktabs}
\usepackage{xcolor}
\usepackage{listings}
\usepackage{amsmath}
\usepackage{amssymb}
\usepackage{mathtools}
\usepackage{amsthm}
\usepackage{bm}
\usepackage{xspace}
\usepackage{enumitem}
\usepackage{arydshln}
\usepackage{lipsum}
\usepackage[capitalize,noabbrev]{cleveref}
\definecolor{upforestgreen}{rgb}{0.6, 0.8, 0.2}
\theoremstyle{plain}
\newtheorem{theorem}{Theorem}[section]

\theoremstyle{definition}

\theoremstyle{remark}

\newcommand\extrafootertext[1]{%
    \bgroup
    \renewcommand\thefootnote{\fnsymbol{footnote}}%
    \renewcommand\thempfootnote{\fnsymbol{mpfootnote}}%
    \footnotetext[0]{#1}%
    \egroup
}

\NewDocumentCommand{\var}{O{s} m O{}}{%
  \ensuremath{#1_{#2}^{#3}}
}
\usepackage{siunitx}

\newcommand{\commentout}[1]{}

\definecolor{light-gray}{gray}{0.80}

\newcommand\fref{Fig.~\ref}
\newcommand\tref{Tab.~\ref}
\newcommand\sref{Sec.~\ref}

\usepackage{amsthm}

\usepackage{amsthm}

\newtheorem{mytheorem}{Theorem}
\newtheorem{assumption}[mytheorem]{Assumption}
\newtheorem{lemma}[mytheorem]{Lemma}
\newtheorem{remk}[mytheorem]{Remark}

\newcommand{\tokenbypass}{TokenBypass\xspace}
\newcommand{\ltd}{LTD\xspace}

\newcommand{\mha}{MHA\xspace}
\newcommand{\ffc}{FFC\xspace}
\newcommand{\bert}{BERT\xspace}
\newcommand{\vit}{ViT\xspace}

\newcommand{\pslg}{MSLG\xspace}

\def\bertbase{BERT$_{\text{base}}$\xspace}
\def\bertlarge{BERT$_{\text{large}}$\xspace}

\newcommand{\gpthf}{GPT-2$_{\text{350M}}$\xspace}
\newcommand{\OURS}{DeepSpeed Data Efficiency\xspace}

\title{\OURS: Improving Deep Learning Model Quality and Training Efficiency via Efficient Data Sampling and Routing}
\author {
    Conglong Li\equalcontrib,
    Zhewei Yao\equalcontrib,
    Xiaoxia Wu\equalcontrib,
    Minjia Zhang,
    Connor Holmes,
    Cheng Li,
    Yuxiong He
}
\affiliations {
    Microsoft
}

\begin{document}

\maketitle
\begin{abstract}
Recent advances on deep learning models come at the price of formidable training cost. The increasing model size is one of the root causes, but another less-emphasized fact is that data scale is actually increasing at a similar speed as model scale, and the training cost is proportional to both of them. Compared to the rapidly evolving model architecture, how to efficiently use the training data (especially for the expensive foundation model pretraining) is both less explored and difficult to realize due to the lack of a convenient framework that focuses on data efficiency capabilities. To this end, we present \OURS, a framework that makes better use of data, increases training efficiency, and improves model quality. Specifically, we propose and combine two data efficiency techniques: efficient data sampling via a general curriculum learning library, and efficient data routing via a novel random layerwise token dropping technique. For GPT-3 1.3B language model pretraining, our work achieves 12.5x less data/time/cost (\$3.7K if rent on Azure), while still maintaining 95\% of model quality compared to baseline with full data and cost (\$46.3K). For GPT-3 1.3B and BERT-large pretraining, our work can also achieve the same model quality with up to 2x less data/time/cost, or achieve better model quality under same data/time/cost. \OURS is easy to use and tune, enabling us to easily apply it and verify its benefit on additional tasks including GPT-3 MoE model pretraining and small-scale GPT-2/ViT finetuning.
\end{abstract}

\section{Introduction}
\label{sec:intro}
\begin{figure}[t]
\centering
\includegraphics[width=0.35\textwidth]{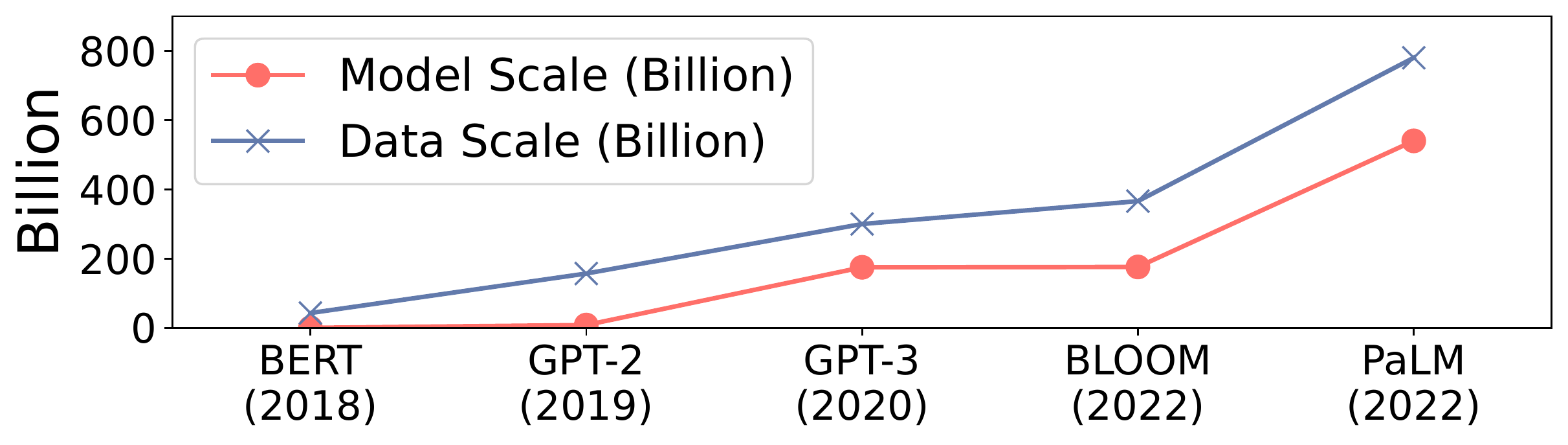}
\caption{Model scale (number of parameters) and data scale (number of consumed training tokens ) of representative language models in the last 5 years~\cite{bert, megatron, gpt3, bloom, palm}.}
\label{fig_model_data_scale}
\end{figure}

Recently, large-scale deep learning models are empowering us to achieve more in many ways, such as code generation~\cite{copilot} and text-to-image generation~\cite{dalle, stablediffusion}. 
To keep improving the service quality, deep learning model architecture evolves rapidly, and the model size is also growing at a tremendous speed. 
The increasing model size leads to unprecedented training cost (especially for foundation model pretraining), which recently grows to 2 months on thousands of GPUs/TPUs~\cite{mt-nlg, palm}. 
On the other hand, a less-emphasized perspective is that \textbf{data scale is actually increasing at a similar speed as model scale, and the training cost is proportional to both of them}. 
As plotted in~\fref{fig_model_data_scale}, for several representative language models in the last 5 years both the model and data scales increase at a similar speed.
Recent works including Chinchilla~\cite{hoffmann2022training} and PaLM 2~\cite{palm2} emphasize the need of increasing data scale at an even faster speed.
This demonstrates the importance of improving data efficiency: achieve same model quality with less data and reduced training cost, or achieve better model quality with the same amount of data and similar training cost.

\extrafootertext{Extended version of this paper (including appendix) can be found on arxiv~\cite{li2022deepspeed}.}

\begin{figure}[t]
\centering
\includegraphics[width=0.3\textwidth]{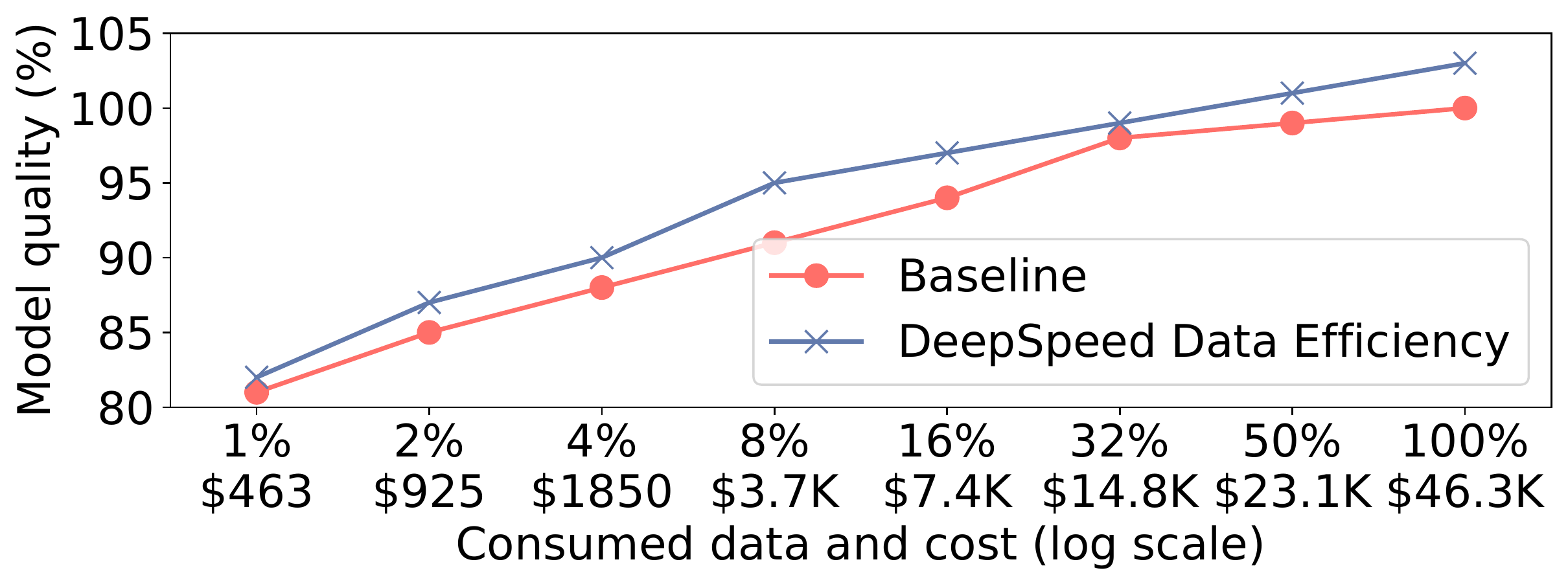}
\caption{GPT-3 1.3B pretraining: relative model quality (baseline with full data as 100\% quality) under different data consumption (1\% to 100\%) and training cost (when renting on Azure).}
\label{fig:moti_pareto}
\end{figure}

There are two popular research directions among existing data efficiency techniques: Data sampling techniques aim to improve the convergence speed by sampling the most suitable next data batch from the whole data pool; Data routing techniques aim to reduce the computation by routing each data to only a subset of the model components. 
These techniques improve data and training efficiency, but existing solutions have several limitations:
\begin{itemize}[noitemsep, nolistsep, labelindent=0pt, leftmargin=*]
\item Techniques like curriculum learning (CL) improve data efficiency by indexing and sampling training data based on certain difficulty metric~\cite{bengio2009curriculum}, and it has recently proved effective on large-scale pretraining tasks~\cite{slw}. 
However, implementing different CL strategies for different user tasks can require a lot of code-refactoring, which is time-consuming and error-prone.
In addition, existing implementations have less consideration on scalability, which makes it difficult to analyze and index large-scale training data based on different difficulty metrics.

\item Existing data routing techniques such as token drop/bypass/pruning were mostly designed for inference and inapplicable to training. TokenBypass~\cite{hou-etal-2022-token}, to our knowledge the only data routing technique for foundation model pretraining, skips the compute of part of the input tokens at some middle layers during BERT pretraining, reducing pretraining cost while maintaining model quality. 
However, it requires several special implementations that may only work for the tested BERT pretraining case, such as the importance score-based token dropping decisions and the whitelist for special tokens. 
This could limit the possibility and benefit of applying it to other cases.

\item Although promising data efficiency solutions have been proposed independently, even a small customization to the strategy would require nontrivial changes in multiple places deep inside the training pipeline: data loader, data sampler, model architecture, etc. Another challenge is that existing techniques usually add additional hyperparameters but without a clear and low-cost tuning strategy.
\end{itemize}
To address these above challenges, we present \OURS, a framework that makes better use of data, increases training efficiency, and improves model quality. 
Specifically, \OURS demonstrates the following contributions:
\begin{itemize}[noitemsep, nolistsep, labelindent=0pt, leftmargin=*]
\item \textbf{Efficient data sampling via general curriculum learning library.} We present a general curriculum learning (CL) library that is both scalable and customizable: it includes a map-reduce based data analyzer that enables scalable analysis and indexing of massive data based on any possible CL metric; it includes a general CL-based data sampler and loader design for users to apply any customized CL strategies. 
Using this library, we are able to thoroughly explore different CL strategies for GPT-3 1.3B and BERT-large pretraining, and identify the best solution that provides better data and training efficiency than existing CL solution. 
This library (and the whole \OURS framework) has been open sourced in a deep learning acceleration library (name hidden for anonymity) that is fully compatible with PyTorch.
This will benefit the whole community as a useful tool to apply curriculum learning to their own training tasks.

\item \textbf{Efficient data routing via random layerwise token dropping.} We present a novel data routing technique called random layerwise token dropping (random-LTD) to skip the computation of a subset of the input tokens at all middle layers. 
Random-LTD employs a simple yet effective routing strategy and requires minimal model architecture change. 
It is very flexible to apply random-LTD to various tasks (GPT-3/GPT-3 MoE/BERT pretraining and GPT/ViT finetuning) which the SOTA technique (\tokenbypass) does not explore or provides less improvement. 

\item \textbf{An easy to use/tune framework that maximizes data/training efficiency.} \OURS seamlessly composes the two proposed techniques, and only requires minimal changes on user side.
We demonstrate that composing data sampling and routing techniques can lead to even better data/training efficiency, especially for foundation model pretraining: For GPT-3 1.3B pretraining, \fref{fig:moti_pareto} shows that our approach provides better model quality at all cost budgets, advancing the whole cost-quality Pareto frontier. In particular, we achieve up to 12.5x data/time/cost saving while still maintaining 95\% of the model quality (zero-shot eval accuracy) compared to the baseline with full data, while baseline can only maintain 91\% of the model quality, a 1.8x higher quality degradation. It requires 2x cost to achieve 95\% quality without our approach.
Based on measured training time, 12.5x would be a cost reduction from \$46.3K to \$3.7K if renting similar hardware on Azure~\cite{azure}, greatly democratizing research and usage of foundation models for AI community.
For GPT-3 1.3B and BERT-large pretraining, we can also achieve up to 2x data and 2x time saving together with better or similar model quality as compared to the baseline training with full data, greatly surpassing state-of-the-art data efficiency solutions as summarized in~\tref{tab_summary}. 
Both techniques under our framework are easy to use and tune, and we include a low-cost tuning strategy and a summarized usage guidelines. 
This enables us to easily apply proposed work and verify its benefits on additional workloads including GPT-3 Mixture-of-Experts (MoE) model pretraining and small-scale GPT-2/ViT model finetuning.
The proposed framework has been open sourced in a deep learning acceleration library called DeepSpeed\footnote{https://github.com/microsoft/DeepSpeed}.
\end{itemize}

\begin{table}[t]
\begin{adjustbox}{width=0.999\linewidth}
\scriptsize
\centering
\begin{tabular}{@{}l@{\hskip 0.02in}|@{\hskip 0.02in}c@{\hskip 0.02in}|@{\hskip 0.02in}c@{\hskip 0.02in}|@{\hskip 0.02in}c@{\hskip 0.02in}|@{\hskip 0.02in}c@{}}
\toprule
& Efficient & Efficient  & Verified & Key \\
 & data sampling & data routing & workloads & achievements \\
\midrule
SLW & 1 specific & N/A & GPT-2/GPT-3 & 1.3x saving with\\
 & CL metric &  & pretraining & 100\% model quality \\
\midrule
\tokenbypass & N/A & \tokenbypass & BERT & 1.33x saving with \\
&  &  & pretraining & 100\% model quality \\
\midrule
Proposed DeepSpeed & general CL & random-LTD & GPT-3/BERT/MoE & 12.5x saving with\\
Data Efficiency & library support &  & pretraining & 95\% model quality\\
 &  &  & GPT-2/ViT & 2x saving with\\
&  &  & finetuning & 100\% model quality\\
\bottomrule
\end{tabular}
\end{adjustbox}
\caption{Comparing \OURS with SOTAs~\cite{slw, hou-etal-2022-token}.}\label{tab_summary}

\end{table}

\section{Background and Related Works}
\label{sec:background}
\textbf{Data sampling.} For deep learning, the most common data sampling method for minibatch stochastic gradient descent is uniform sampling, where at each step a batch of data is drawn uniformly at random from the whole training data. 
However, it's potentially beneficial to focus on different kinds of data at different training stages. 
One example is the curriculum learning technique~\cite{bengio2009curriculum} which aims to improve training convergence speed by presenting relatively easier or simpler examples earlier during training. 
Building a curriculum learning solution usually requires two components: the difficulty metric (i.e., how to quantify the difficulty of each data sample) and the pacing function (i.e., how to decide the difficulty range when sampling next training data batch). 
In the NLP area, curriculum learning has been applied on small-scale one-stage tasks and downstream finetuning tasks, such as neural machine translation (NMT)~\cite{kocmi2017curriculum, bojar2017results, zhang2018empirical, platanios2019competence, zhang2019curriculum} and natural language understanding (NLU)~\cite{sachan2016easy, sachan2018self, tay2019simple, xu2020curriculum}. 
There are also a few works that explore curriculum learning for language model pretraining~\cite{press2020shortformer, zhang2021reducing, campos2021curriculum, slw}. 
However, one common limitation among existing works is that there does not exist a scalable and customizable curriculum learning library, making it difficult to analyze large-scale data and explore custom difficulty metrics/pacing functions. 
One evidence is that most of the curriculum learning works for language model pretraining only focus on the sequence length metric due to the difficulty of exploring other metrics on the huge pretraining dataset.

\textbf{Data routing.} In common deep learning training, the model is considered as a whole and all sampled data will be routed to all model components. 
However, it's potentially beneficial to route each data sample to only a subset of model components, improving the training efficiency. 
One direction of efficient data routing is to add data bypassing/skipping capability to existing model architectures such as Transformer. 
Transformer~\citep{vaswani2017attention} architecture is a stack of transformer layers, each of which has two main ingredients, i.e., the multi-head attention (\mha) and the feed-forward connection network (\ffc). 
Suppose the transformer has $l$ layers denoted as $L_1,\ldots,L_{l}$.
Let $X_{i} \in \mathbb{R}^{s\times d}$ be the output tensor of $i-$th transformer layer, and $x_0$ be the input (after embedding) of the transformer. 
Here $s$ is the sequence length and $d$ is the hidden dimension. 

Several token dropping/bypassing/pruning techniques~\cite{kim2021learned,goyal2020power,kim2020length,press2021train,wang2021spatten} were proposed for \bert inference to reduce the computational overhead, but they are not practical for training.
In these works, if a token $i$ ($X_{j, i}$) is decided to be dropped at layer $j$ ($L_j$), the compute cost of this token through all remaining layers ($L_k$ where $k>j$) is eliminated. 
As such, the sequence length $s_i$ of the $i$-th layer's input $X_{i-1}$ will be a non-increasing array, i.e., $s_0 \geq s_1~...~\geq s_{l}$. 
However, such a configuration has been shown instability for adaptive token-dropping inference~\citep{kim2020length}.
Therefore, \cite{kim2020length} utilize the sandwich rule and  distillation from~\citep{yu2019universally} to stabilize training and boost accuracy.
But these two methods also significantly increase the training cost. 
Thus, such techniques cannot be applied to speed up the pretraining procedure.

Recently, \tokenbypass~\cite{hou-etal-2022-token} enabled token dropping for \bert pretraining. 
It uses several importance scores/metrics to determine the dropped tokens (token frequency and cumulative loss). 
It proposed two main mechanisms to overcome the training instability issue: 
(1) the sandwich token dropping rule, where the first ($L_1$ to $L_i$) and the last few \bert layers ($L_{l-j}$ to $L_l$) capture all tokens (no token dropping) and only bypass $s' \leq s$ tokens from $L_i$ to $L_{l-j}$ middle layers.
Particularly, the authors (only) test on the encoder transformer (12-layer \bertbase and 24-layer \bertlarge), and let $i=l/2-1$, $j=1$, $s'=s/2$.
(2) special token treatment, where special tokens (e.g., \texttt{[MASK], [CLS], [SEP]}) are never dropped. 
Compared to \tokenbypass, our random-LTD 
(1) does not require importance score metric, special token treatment, or the sandwich token dropping rule, which dramatically reduces the manual design effort;
(2) has been broadly tested on GPT-3/BERT pretraining tasks and GPT-2/ViT finetuning tasks, providing better data/training efficiency than \tokenbypass.

\begin{figure}[t]
\centering
\includegraphics[width=0.25\textwidth]{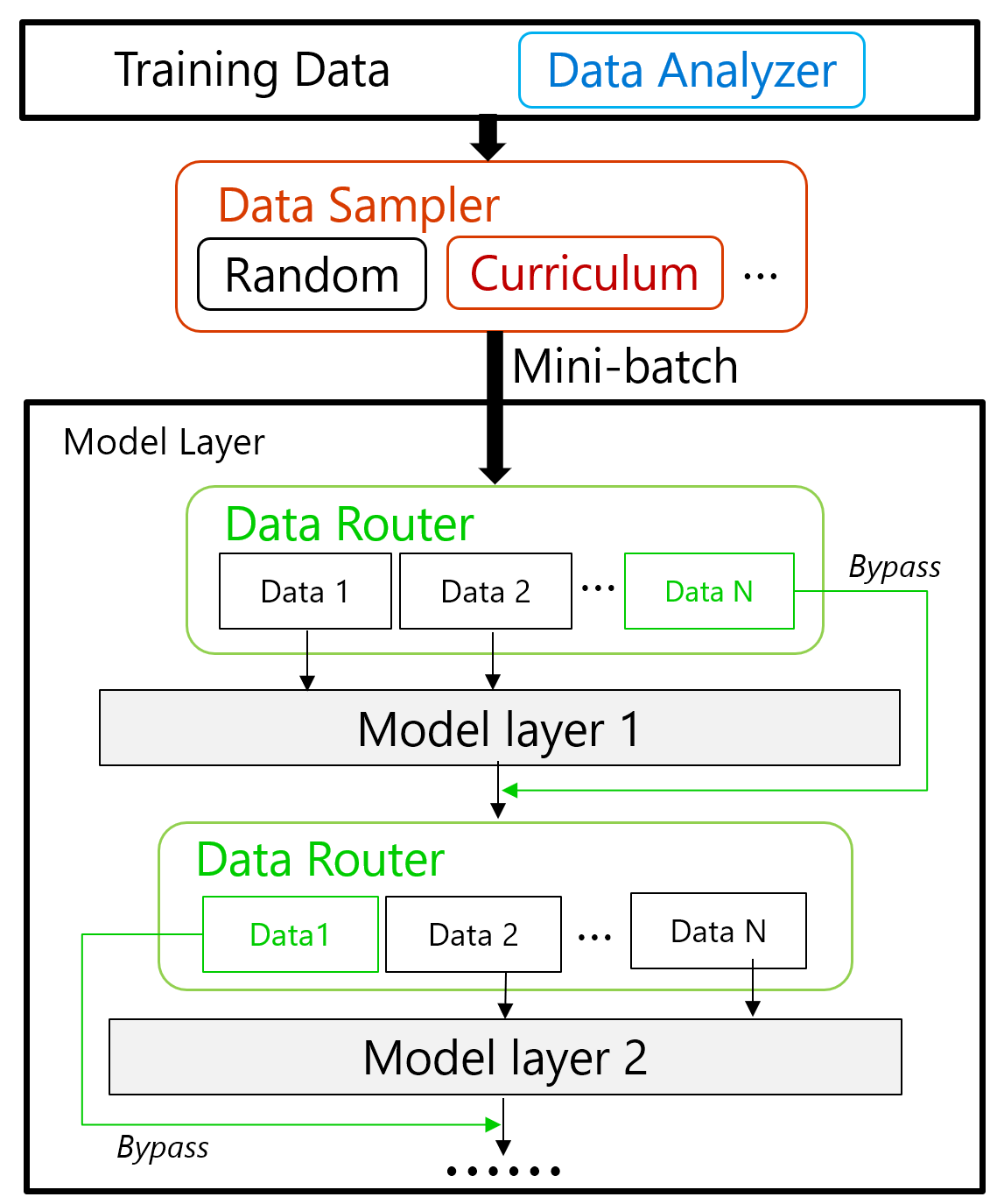}
\caption{Design of the \OURS framework.}
\label{fig_design_data_efficiency}
\end{figure}
\section{Design}
\label{sec:design}
At high-level, the proposed \OURS framework has two components as shown in~\fref{fig_design_data_efficiency}: 
First we have efficient data sampling, where instead of the baseline’s random sampling, we aim to sample the most suitable next data batch from the whole data pool by a general curriculum learning (CL) library.
Second we have efficient data routing, where instead of passing all input data to all model components, we aim to efficiently route each data through different components of model by leveraging the proposed random layerwise token dropping (random-LTD) technique. 
This section presents the design of the two techniques, how we compose them, together with a low-cost tuning strategy and a summarized usage guidelines.

\subsection{Efficient data sampling via curriculum learning}
\label{sec:design-cl}
To solve the limitations of existing CL solutions as described in previous sections, we design and implement a general curriculum learning library emphasizing the scalability and customizability. 
It consists of three components as shown in top part of~\fref{fig_design_data_efficiency}. 
First we use a data analyzer to perform the offline CPU-only data analysis which indexes the whole data pool based on any difficulty metric, which could be the sequence length, the vocabulary rarity, or anything defined by user. 
This data analyzer employs a Map-Reduce scheme: During the Map stage, user provides a function that computes the desired difficulty metric, the raw training dataset, and other configurations such as number of CPU nodes and number of threads per node. 
Then the data analyzer will automatically splits the dataset based on number of workers, compute the difficulty values in a batched fashion, and write the results to two indexes: one index maps each data sample to its difficulty value, and another index maps each distinct difficulty value to the corresponding samples. 
During the Reduce stage, the data analyzer will merge the index files produced by all workers. 
This Map-Reduce scheme is necessary since the training data could be huge thus has to be distributed.
For instance, we have 173 million data samples (each with sequence length 2048) for GPT-3 pretraining and 2.5 billion data samples (each with sequence length $\leqslant$ 512) for BERT pretraining. 
To reduce the memory overhead when analyzing the huge dataset, we write the index files as numpy memory-mapped files. 
Using this data analyzer we are able to efficiently analyze GPT-3 and BERT pretraining data based on various difficulty metrics. 
Using 40 CPU threads on a single node with AMD EPYC 7V12 64-Core Processor, we can finish the analysis on one metric within 3/80 hours for GPT-3/BERT data, respectively.

Next, during training, the curriculum scheduler will determine the difficulty threshold for the current step based on a pacing function such as linear, rooted, or any strategy provided by user. 
Then the data sampler will sample the data with desired difficulty from the indexed data pool. 
To apply the proposed CL solution to a existing training pipeline, user just need to call an API and provide the raw training data, the difficulty metric index (computed in the offline analysis), and the pacing function configurations. 
Our framework will then provide a curriculum learing-based data loader that users can simply iterate at each step. 
Using our CL library for GPT-3/BERT pretraining, we are able to easily analyze and index the huge training data based on 7 difficulty metrics:
\begin{itemize}[noitemsep, nolistsep, labelindent=0pt, leftmargin=*]
\item \textbf{Truncation-based sequence length (seqtru), for GPT and BERT.} This metric starts with shorter data samples and gradually increases the sequence length during training. 
To change the sequence length, this metric will truncate the sequences (from the end of sequence) while keeping the number of samples unchanged, thus the number of tokens will decrease. 
This metric is recently applied to GPT-2 and GPT-3 models and demonstrate decent training efficiency gains~\cite{slw}. 

\item \textbf{Reshape-based sequence length (seqres), for GPT.} This metric is similar to seqtru metric, but instead of truncating we break the original sequences into segments based on the desired new sequence length. 
Thus we are essentially ``reshaping'' the input tensor into more samples and shorter lengths. 
This metric is proposed in MosaicML Composer as a variant of the seqtru metric~\cite{mosaicslw}, but their documentation does not describe which way provides better model quality. 
We don't apply the seqres to BERT case because unlike GPT data where all tokens are valid, BERT input sequences only include two natural sentences thus each sequence has different ``effective sequence length'' and then padded to 512. 
If we simply ``reshape'' BERT sequences, some of the new short sequences may only contain padding tokens.

\item \textbf{Reorder-based sequence length (seqreo), for BERT.} This metric is similar to seqtru metric, but instead of truncating we adjust the sequence length by reordering the training data based on the ``effective sequence length'' in BERT training data sequences.

\item \textbf{Vocabulary rarity (voc), for GPT and BERT.} This metric was proposed in a CL work for neural machine translation~\cite{platanios2019competence}. 
It computes the product of the unigram probabilities for each sequence by $-\sum_{k=1}^{N} log(p(w_k))$ where $p(w_k)$ is the vocabulary frequency (inside whole training data) of the $k$th word in the sequence. 
Lower value indicates that the sequence has more common vocabularies. 

\item \textbf{seqtru\_voc, for GPT and BERT. seqres\_voc, for GPT. seqreo\_voc, for BERT.} These 3 metrics are combinations of above metrics. 
For seqtru\_voc and seqres\_voc, we first reorder the training data based on voc metric, then apply seqtru or seqres as a kind of post-processing. 
For seqreo\_voc, we treat it as a single new metric and index the data based on it.
\end{itemize}
Besides the difficulty metrics, another set of CL hyperparameters is the pacing function: the start and end difficulty ($d_s$ and $d_e$), total number of CL steps ($T_{c}$), and the kind of pacing function (linear, sqrt, or users can plug in any customized function to the proposed framework). 
For seqtru and seqres metrics, we set the $d_s$ and $d_e$ as value-based (e.g., $d_s=80$, $d_e=2048$) since the possible values of these two metrics are continuous. 
For other metrics, we set $d_s$ and $d_e$ as percentile-based (e.g., $d_s=1\%$, $d_e=100\%$) since the possible values of these metrics are discrete. 
For seqtru and seqres we use a linear pacing function ($d_t = d_s + (d_e - d_s) \times min(\frac{t}{T_{c}}, 1)$) following the preivous work~\cite{slw}, while for seqreo and voc we use a sqrt pacing function ($d_t = d_s + (d_e - d_s) \times min((\frac{t}{T_{c}})^{0.5}, 1)$). 
This is because seqreo and voc will only sample from a subset of data pool before reaching the end difficulty, and previous work finds that in such case it's beneficial to use a sqrt function to avoid sampling too much easy samples at the beginning~\cite{platanios2019competence}. \sref{sec:design-compose} includes low-cost tuning strategy and usage guidelines for our CL solutions.

\subsection{Efficient data routing via random-LTD}
\label{sec:design-ltd}
\textbf{Layerwise Token Dropping.} Existing token dropping methods for inference and training either permanently drop tokens from the compute graph at intermediate layers, or at least make some tokens fully skip a consecutive series of middle layers (\sref{sec:background}). 
However, several works~\citep{vig2019analyzing,michel2019sixteen,voita2019analyzing} have shown that \mha focuses on different tokens at different layer depths and the attention map aligns with the dependency relation most strongly in the middle of transformer architectures. 
Therefore, fully skipping middle layers like \tokenbypass~\cite{hou-etal-2022-token} may hinder the learnability/generalization of the architecture during pretraining/inference. 
We conjecture that this might be why multiple first/last layers need to disable token bypassing and the special token treatment is needed. 

In order to overcome this problem, we propose a layerwise token dropping (\ltd) mechanism. 
Instead of fully bypassing same tokens over all middle layers, each transformer layer independently drops/retains its own set of tokens.
In more detail, recall that the input of $(i+1)$-th layer ($L_{i+1}$) is $X_{i}\in\mathbb{R}^{s\times d}$. 
Denote the dropped token index as $J_i=\{j_1, j_2, ..., j_{a_i}\}$ and the kept token index as $K_i=\{k_1, ..., k_{b_i}\}$ such that $a_i+b_i=s$.
We have $J_i\cup K_i = \{1,2,3...,s\}$ and $J_i\cap K_i=\emptyset$ for each layer.
Meanwhile, for any two different layers $L_{i_1}$ and $L_{i_2}$, $J_{i_1}$ and $J_{i_2}$ are independent, though the dropped ratios are the same.
With this layerwise mechanism, each token rarely bypasses all middle layers. Thus, its dependency on other tokens can be captured by \mha. 

\textbf{Random Token Dropping.}
Various importance score-based metrics are used to determine the token dropping criterion. 
Most of them can be categorized in attention score-based or loss/frequency-based metrics. 
However, both of them introduce challenges that make LTD less practical: 
For attention score-based metrics, the compute cost for \ltd is too high since the metric has to be calculated for every layer; 
For loss/frequency-based metrics, the accumulated loss or frequency would not be changed within the same iteration, which leads the dropped token to be the same for different layers, breaking the desired \ltd mechanism. 
Instead of importance score, we propose to use \emph{purely random} token dropping assignment and prove its effectiveness in all our experiments.
For each transformer layer, we randomly (uniformly) select a small batch of tokens to proceed with the compute and drop the rest. 
In more details, assume $M_i=$\{$m_i(1)$, $m_i(2)$, ..., $m_i(s)$\} is a random shuffle of $S=$\{1, 2, ..., s\}. 
Then the dropped token set is $J_i=$\{$m_i(1)$, $m_i(2)$, ..., $m_i(a_i)$\} for the input of $L_{i+1}$.

\begin{figure}[t]
\centering
\includegraphics[width=0.3\textwidth]{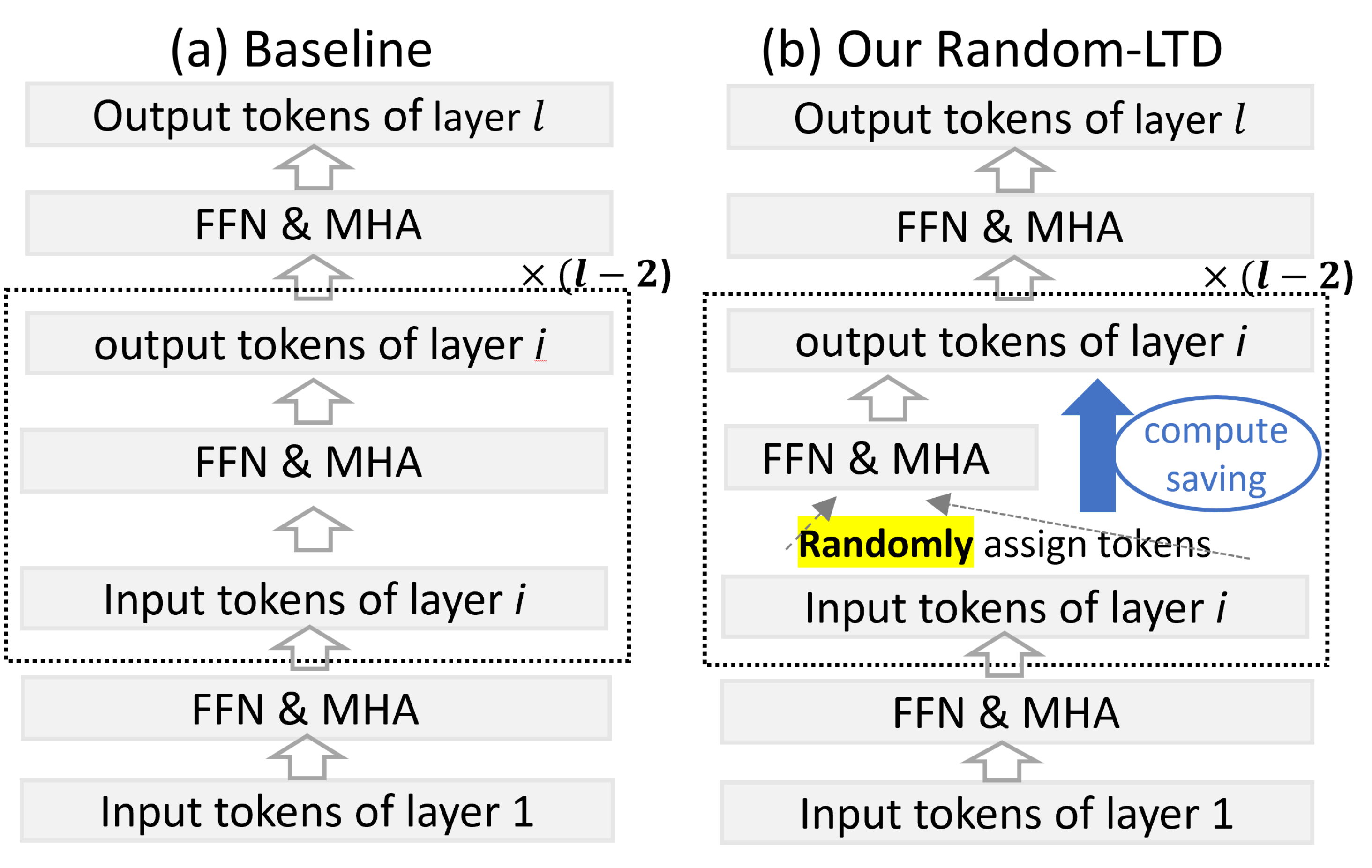}
\caption{Transformer layers for baseline and random-LTD. The dash-line box is repeated by $l-2$ times.}
\label{fig:illustration_of_random_ltd_and_baseline}
\end{figure}

\textbf{Random and Layerwise Token Dropping.}
Combining layerwise token dropping with random token dropping, we have our final random and layerwise token dropping method (random-LTD), which can efficiently apply token dropping for each individual layer and can capture the attention dependency of each token with other others in middle layers with high probability. 
As a result, our experiments on BERT pretraining confirm that random-LTD does not require and won't benefit from special token treatment used by the \tokenbypass work, further reducing the implementation complexity. 
\fref{fig:illustration_of_random_ltd_and_baseline} presents the comparison between standard baseline training and random-LTD.  
For each layer, random-LTD randomly selects (function ``gather'') a subset of the tokens and feeds (function ``Layer'') them into the transformer layer. 
Afterward, we combine (function ``combine'') the output of transformer layer with the dropped tokens to recover the full sequence length in a order-preserved manner. 
Thus, the next layer still receives the full sequence and can repeat this process. 
To apply random-LTD to an existing training pipeline, user just needs to provide the module class name that they want to apply random-LTD (e.g., a TransformerLayer class). 
Then \OURS will wrap the module with a new module that includes token dropping capability, and drop some of the input tokens for this module during training.


\textbf{Layers without Token Dropping.}
While \tokenbypass needs to keep half of the layers in full sequence length training, random-LTD has no such limitation. 
Thanks to its attention-capture feature, we can apply random-LTD to most of the transformer layers except the first and last layers, enabling further training efficiency gain. 
Our experiments show that keeping the first and last layers in full sequence length training usually leads to better performance since 
(1) the first layer directly connects to the embedding, and it can help refine the raw feature;
(2) directly connected to the final prediction, the last layer provides a feature realignment for all tokens which could improve the model quality. 


\textbf{Monotonic Sequence Length Growth.}
%
In order to reduce the gradient variance introduced by random-LTD, we gradually increase the kept sequence length throughout training with a linear schedule (referred to as \pslg). 
Thus random-LTD has two hyperparameters similar to CL: starting from a sequence length $r_s$ which denotes the size of kept token set $K_i$ for each middle layer after dropping, random-LTD will gradually drop less tokens (following a linear function) and eventually stop dropping after $T_{r}$ steps. 
Our experiments show that \pslg provides better model quality than constant drop schedule under similar data/compute savings.
\sref{sec:design-compose} includes low-cost tuning strategy and usage guidelines for random-LTD.


\begin{table}[t]
  \scriptsize
  \centering
  \begin{tabular}{@{}l@{\hskip 0.01in}|@{\hskip 0.01in}l@{}}
  \toprule
Case & Guidelines \\
\midrule
GPT-3  & CL: $d_s$ = 80/1\% (seqtru/voc), $T_{c}$ = 40\% of baseline's total steps \\
pretraining & random-LTD: $r_s$ = 128, $T_{r}$ = 70\% of baseline's total steps \\
\midrule
BERT  & CL: $d_s$ = 128/5\% (seqtru/voc), $T_{c}$ = 50\% of baseline's total steps \\
pretraining & random-LTD: $r_s$ = 128, $T_{r}$ = 100\% of baseline's total steps \\
\midrule
GPT-2  & CL: $d_s$ = 32 (seqres), $T_{c}$ = 70\% of baseline's total steps \\
finetuning & random-LTD: $r_s$ = 128, $T_{r}$ = 30\% of baseline's total steps \\
\midrule
ViT finetuning & random-LTD: $r_s$ = 32/66, $T_{r}$ = 80\% of baseline's total steps \\
\bottomrule
  \end{tabular}
  \caption{CL and random-LTD usage guidelines.}\label{table_guide}
\end{table}
\subsection{Composing CL and random-LTD, tuning strategy, usage guidelines}
\label{sec:design-compose}
CL and random-LTD are complementary: CL helps to sample the next data batch, and random-LTD helps to decide how to route each sampled data inside the model. 
\OURS hides several complexities when composing the two techniques so that users can easily enjoy the compound benefit. 
As one example, some CL metrics would affect the actual sample sequence length, thus inside our framework we make sure the random-LTD's token dropping mechanism is aware of this, and also adjust the calculation of number of actual consumed tokens which are affected by both techniques. 
This token consumption calculation is also critical to the learning rate schedule: previous CL work~\cite{slw} finds that if a CL technique reduces the number of tokens on certain steps, it is desirable to use a learning rate decay schedule based on consumed tokens instead of consumed steps. This is because if baseline and CL use the same step-wise LR decay, it leads to much faster token-wise LR decay for CL which hurts model quality. In this work, we apply the token-based LR decay schedule for both CL and random-LTD. To our knowledge this is the first work to apply such LR schedule to token dropping/data routing techniques, and our experiments show that it does help improving random-LTD's performance. Our CL library's general data analyzer/sampler/loader and random-LTD's module wrapping design makes it easy to apply our framework to different model training tasks. And the overall composibility of \OURS enables us to leverage both data efficiency techniques and achieve even better data and training efficiency (\sref{sec:eval}).

\textbf{Tuning Strategy and Usage Guidelines.} Both CL and random-LTD only have two parameters that need user tuning: the starting CL difficulty/random-LTD seqlen ($d_s$/$r_s$), and the total CL/random-LTD steps ($T_{c}$/$T_{r}$).~\footnote{For CL, the ending difficulty $d_e$ is always the highest possible difficulty.} And for both CL and random-LTD we find that it's possible to apply a low-cost tuning strategy proposed in previous CL work~\cite{slw}, where we perform binary search on a very small portion (e.g., 2\%) of training to find the smallest $d_s$/$r_s$ and largest $T_{c}$/$T_{r}$ that don't trigger substantial validation loss fluctuations (``whether the perplexity value becomes larger than 1.3x of the previous best perplexity''). For GPT-2 finetuning, given the low training cost we also perform full training of 16 different CL/random-LTD settings which confirm that (1) the low-cost tuning strategy is able to find very good hyperparameters; (2) both CL and random-LTD are not sensitive to hyperparameter choices. \tref{table_guide} summarizes the usage guidelines based on our tuning results, which we believe can be directly applied to any similar models (at least as a very good starting point for any further tuning).

\section{Evaluation}
\label{sec:eval}
We evaluate \OURS by GPT-3/GPT-3 MoE/BERT pretraining and GPT-2/ViT finetuning. 
Appendix A.5 includes studies of the \tokenbypass method on GPT finetuning and pretraining, further demonstrating the advantages of the proposed random-LTD method.

\subsection{GPT-3 and GPT-3 MoE pretraining}
\label{sec:eval-gpt}
We use \textit{the Pile} public dataset~\cite{gao2020pile} to perform the pretraining of GPT-3 1.3B~\cite{gpt3} (24 layers, 2048 hidden size, 16 attention heads) model. We also pretrain a GPT-3 Mixture-of-Experts (MoE) 6.7B model (24 layers, 1024 hidden size, 16 attention heads, 64 experts on every other layer) following related work~\cite{rajbhandari2022deepspeed}.
We then perform 0-shot and 10-shot evaluations on 19 tasks to evaluate the model quality of the pretrained models. 
Detailed experimental setup and additional discussions on results are described in Appendix A.1.

\begin{table}[t]
\begin{adjustbox}{width=0.999\linewidth}
\scriptsize
\begin{tabular}{@{}l@{\hskip 0.01in}|@{\hskip 0.01in}c@{\hskip 0.01in}|@{\hskip 0.01in}c@{\hskip 0.01in}|@{\hskip 0.01in}c@{\hskip 0.01in}|@{\hskip 0.01in}c@{\hskip 0.01in}|@{\hskip 0.01in}c@{}}
\toprule
& CL/& Data & Time  & Avg & Avg \\
& random-LTD& (billon & (hours on  & 0-shot & 10-shot \\
Case & hyperparameter& tokens) & 64 V100) & accuracy & accuracy\\
\midrule
(1)baseline & N/A & 300 (1x) & 260 (1x) & 42.5 & 44.0 \\
(2)CL\_seqtru & $d_s=80$, $T_{c}=110K$ & 300 (1x) & 257 (1.01x) & 43.4 & 44.8 \\
(3)CL\_seqres & $d_s=80$, $T_{c}=110K$ & 300 (1x) & 248 (1.05x) & 43.0 & 44.5 \\
(4)CL\_voc & $d_s=1\%$, $T_{c}=110K$ & 300 (1x) & 257 (1.01x) & 42.3 & 44.5 \\
(5)CL\_seqtru\_voc & same as (2) + (4) & 300 (1x) & 259 (1.00x) & 43.6 & 44.9 \\
(6)CL\_seqres\_voc & same as (3) + (4) & 300 (1x) & 248 (1.05x) & 43.0 & 44.4 \\
(7)random-LTD & $r_s=128$, $T_{r}=200K$ & 300 (1x) & 263 (0.99x) & 43.7 & 44.9 \\
{\textbf{(8)CL\_seqtru\_voc}} & same as (5) + (7) & 300 (1x) & 260 (1.00x) & {\textbf{43.8}} & {\textbf{45.1}} \\
{\textbf{+random-LTD}} & &  &  &  &  \\
\midrule
(9)baseline & N/A & 200 (1.5x) & 174 (1.49x) & 41.9 & 44.0 \\
(10)CL\_seqtru\_voc & seqtru: $d_s=80$, $T_{c}=73K$ & 200 (1.5x) & 171 (1.52x) & 42.7 & 44.5 \\
& voc: $d_s=1\%$, $T_{c}=73K$ &  &  &  &  \\
(11)random-LTD & $r_s=128$, $T_{r}=133K$ & 200 (1.5x) & 176 (1.48x) & 43.1 & 44.8 \\
\midrule
(12)baseline & N/A & 150 (2x) & 130 (2.00x) & 42.0 & 42.7 \\
(13)CL\_seqtru\_voc & seqtru: $d_s=80$, $T_{c}=55K$ & 150 (2x) & 129 (2.02x) & 42.6 & 43.7 \\
& voc: $d_s=1\%$, $T_{c}=55K$ &  &  &  &  \\
(14)random-LTD & $r_s=128$, $T_{r}=100K$ & 150 (2x) & 131 (1.98x) & 42.7 & 43.5 \\
{\textbf{(15)CL\_seqtru\_voc}} & same as (13) + (14) & {\textbf{150 (2x)}} & {\textbf{130 (2.00x)}} & 42.8 & 44.0 \\
{\textbf{+random-LTD}} & &  &  &  &  \\
\midrule
(16)baseline & N/A & 300 (1x) & 111 (1x) & 42.8 &  \\
{\textbf{(17)CL\_seqtru\_voc}} & same as (5) + (7) but with & 300 (1x) & 111 (1.00x) & {\textbf{43.5}} & \\
{\textbf{+random-LTD}} & 2x $T_{c}$ and $T_{r}$ due to batch size &  &  &  &  \\
\bottomrule
\end{tabular}
\end{adjustbox}
\caption{GPT-3 1.3B (case 1 to 15) and GPT-3 MoE 6.7B (case 16, 17) pretraining cost and average evaluation accuracy on 19 tasks. GPT-3 MoE only has 0-shot accuracy due to time constraints. Accuracy results for each single task can be found in Appendix A.1.}\label{tab_gpt_eval}
\end{table}

\tref{tab_gpt_eval} summarizes the evaluation results. Key findings include:
\begin{itemize}[noitemsep, nolistsep, labelindent=0pt, leftmargin=*]
\item Results under 100\% data shows that the CL difficultiy metric (5)CL\_seqtru\_voc provides the best model quality, better than both baseline and the CL metric (2)CL\_seqtru proposed in previous work~\cite{slw} (\tref{tab_gpt_eval} case 1 to 6).
\item With 67\% data, our CL solution is able to achieve better 0-shot and 10-shot accuracy than baseline with 100\% data, achieving a 1.5x data and time saving (case 1, 9, 10).
\item When applying the proposed random-LTD technique, results show similar benefit as CL: better model quality when using 100\% data (case 7), and 1.5x data/time saving while maintaining model quality (case 11).
\item With 100\% data, results (case 5, 7, 8) show that using both techniques together further improves the model quality.
\item With 50\% data, the composed solution is able to achieve the same model quality as baseline with 100\% data, demonstrating a 2x data and 2x time saving (case 15).
\item As presented in~\sref{sec:intro} and~\fref{fig:moti_pareto}, our approach provides better model quality at all cost budgets, advancing the whole cost-quality Pareto frontier. In particular, we achieve up to 12.5x data/time/cost saving (from \$46.3K to \$3.7K if renting similar hardware on Azure~\cite{azure}) while still maintaining 95\% of the model quality compared to the baseline with full data. It requires 2x cost to achieve 95\% quality without our approach.
\item Our approach can also improve MoE model's model quality (case 16, 17), confirming its broad applicability.
\end{itemize}

\begin{table}[t]
\begin{adjustbox}{width=0.999\linewidth}
\scriptsize
\begin{tabular}{@{}l@{\hskip 0.01in}|@{\hskip 0.01in}c@{\hskip 0.01in}|@{\hskip 0.01in}c@{\hskip 0.01in}|@{\hskip 0.01in}c@{\hskip 0.01in}|@{\hskip 0.01in}c@{}}
\toprule
& CL/& Data & Time  & GLUE \\
& random-LTD& (billon & (hours on  & finetune \\
Case & hyperparameter& tokens) & 64 V100) & score\\
\midrule
(1)baseline & N/A & 1049 (1x) & 261 (1x) & 87.29±0.53  \\
(2)CL\_seqtru & $d_s=128$, $T_{c}=960K$ & 1049 (1x) & 265 (0.98x) & 87.31±0.57  \\
(3)CL\_seqreo & $d_s=5\%$, $T_{c}=960K$ & 1049 (1x) & 261 (1.00x) & 87.48±0.61  \\
(4)CL\_voc & $d_s=5\%$, $T_{c}=960K$ & 1049 (1x) & 261 (1.00x) & 87.36±0.64  \\
(5)CL\_seqtru\_voc & same as (2) + (4) & 1049 (1x) & 266 (0.98x) & 87.60±0.34  \\
(6)CL\_seqreo\_voc & same as (3) + (4) & 1049 (1x) & 262 (1.00x) & 87.06±0.52  \\
{\textbf{(7)random-LTD}} & $r_s=128$, $T_{r}=2M$ & 1049 (1x) & 302 (0.86x) & {\textbf{88.17±0.48}} \\
(8)CL\_seqtru\_voc & same as (5) + (7) & 1049 (1x) & 290 (0.90x) & 87.69±0.32 \\
+random-LTD & &  &  &    \\
\midrule
(9)baseline & N/A & 703 (1.5x) & 175 (1.49x) & 87.19±0.49 \\
(10)CL\_seqtru\_voc & seqtru: $d_s=128$, $T_{c}=640K$ & 703 (1.5x) & 178 (1.47x) & 87.29±0.62 \\
& voc: $d_s=5\%$, $T_{c}=640K$ &  &  &    \\
(11)random-LTD & $r_s=128$, $T_{r}=1.34M$ & 703 (1.5x) & 201 (1.3x) & 87.99±0.38 \\
\midrule
(12)baseline & N/A & 524 (2x) & 131 (1.99x) & 86.61±0.5 \\
(13)CL\_seqtru\_voc & seqtru: $d_s=128$, $T_{c}=480K$ & 524 (2x) & 133 (1.96x) & 86.9±0.33 \\
& voc: $d_s=5\%$, $T_{c}=480K$ &  &  &  \\
(14)random-LTD & $r_s=128$, $T_{r}=1M$ & 524 (2x) & 150 (1.74x) & 87.32±0.48 \\
{\textbf{(15)CL\_seqtru\_voc}} & same as (13) + (14) & {\textbf{524 (2x)}} & {\textbf{144 (1.81x)}} & 87.44±0.46 \\
{\textbf{+random-LTD}} & &  &  &  \\
\bottomrule
\end{tabular}
\end{adjustbox}
\caption{BERT-large pretraining cost and GLUE finetuning score (median±std, details in Appendix A.2).}\label{tab_bert_finetune}
\end{table}

\subsection{BERT-large pretraining}
\label{sec:eval-bert}
We use \textit{the Pile} public dataset~\cite{gao2020pile} to perform the pretraining of BERT-large~\cite{bert} (24 layers, 1024 hidden size, 16 attention heads) model. 
We then perform GLUE finetuning to evaluate the model quality of the pretrained models. 
Detailed experimental setup and additional discussions on results are described in Appendix A.2.

\tref{tab_bert_finetune} summarizes the evaluation results. Key findings include:
\begin{itemize}[noitemsep, nolistsep, labelindent=0pt, leftmargin=*]
\item Same as the GPT-3 case, results under 100\% data shows that the CL difficultiy metric (5)CL\_seqtru\_voc provides the best model quality (\tref{tab_bert_finetune} case 1 to 6).
\item With 67\% data, our CL solution is able to achieve on-par GLUE score as baseline with 100\% data, achieving a 1.5x data and time saving (case 1, 9, 10).
\item Random-LTD is able to achieve better GLUE score even with 2x less data than baseline (case 14), greatly surpassing the 1.33x data saving by the state-of-the-art \tokenbypass method. The time saving is less than data saving because the token dropping mechanism adds a computation overhead at each step. However, even with this overhead random-LTD is still a more data/time-efficient solution than baseline and \tokenbypass.
\item At 50\% data, the composed solution further improves the GLUE score from the CL/random-LTD-only cases (case 15), achieving a 2x data and 1.8x time saving while maintaining the GLUE score compared to baseline.
\item At 100\% data, the composed solution (case 8) improves the GLUE score from the CL-only case, but is worse than the random-LTD-only case.
\end{itemize}

\begin{table}[t]
 \centering
		\scriptsize
\begin{tabular}{@{}l@{\hskip 0.01in}|@{\hskip 0.01in}c@{\hskip 0.01in}|@{\hskip 0.01in}c@{\hskip 0.01in}|@{\hskip 0.01in}c@{}}
\toprule
 & Best PPL & Num. combinations & PPL median/std\\
Case & at seed 1234 & surpass baseline & over 5 seeds\\
\midrule
(1)baseline & 16.077 & N/A & 16.077±0.028 \\
(2)CL\_seqtru & 15.888 & 9 out of 16 &  \\
{\textbf{(3)CL\_seqres}} & 15.795 & 16 out of 16 & {\textbf{15.818±0.032}}  \\
(4)CL\_voc & 16.031 & 4 out of 16 &   \\
(5)CL\_seqtru\_voc & 16.005 & 3 out of 16 &   \\
(6)CL\_seqres\_voc & 15.981 & 8 out of 16 &  \\
(7)random-LTD & 15.910 & 16 out of 16 & 15.948±0.040 \\
(8)CL\_seqres & 15.831 & N/A & 15.831±0.014 \\
+random-LTD & &  &     \\
\bottomrule
\end{tabular}
\caption{GPT-2 finetuning on PTB results.}\label{tab_gpt_finetune}
\end{table}

\subsection{GPT-2 and ViT finetuning}
\label{sec:eval-ptb}
To verify the effectiveness of the proposed work on small-scale tasks, we apply our techniques to PTB finetuning task~\cite{marcus-etal-1993-building} for an already-pretrained \gpthf model checkpoint from Huggingface. 
Given the much smaller training cost, we focus on improving the model quality under the same amount of data, and results in \tref{tab_gpt_finetune} shows that the proposed approaches are able to improve the model quality as expected.
Detailed experimental setup, hyperparameter tuning, and additional discussions on results are described in Appendix A.3.

We also try finetune the vision transformer (\vit) on both ImageNet and CIFAR. 
We only test random-LTD for this task, due to time/resource limitation. 
Detailed experimental setup is described in Appendix A.4. 
Results show that random-LTD is able to achieve 1.3-1.4x data savings while maintaining the model quality, demonstrating its broad applicability.

\section{Conclusion}
In this work we propose the \OURS framework, which demonstrates the power of composing 2 novel data efficiency techniques together. This enables us to achieve up to 12.5x data/time/cost saving (from \$46.3K to \$3.7K on Azure) while maintaining 95\% of model quality for GPT-3 pretraining, an up to 2x saving for GPT-3 and BERT pretraining while maintaining 100\% model quality, or to achieve even better model quality under similar data and cost. \OURS is easy to use and tune, which enables us to apply it and verify the benefit on additional GPT-3 MoE pretraining and GPT-2/ViT finetuning tasks.


\bibliography{reference}

\clearpage
\appendix

\section{Appendix}

\begin{figure*}[t]
\centering
\subfigure[Begining 10\% of training]{\includegraphics[width=0.4\textwidth]{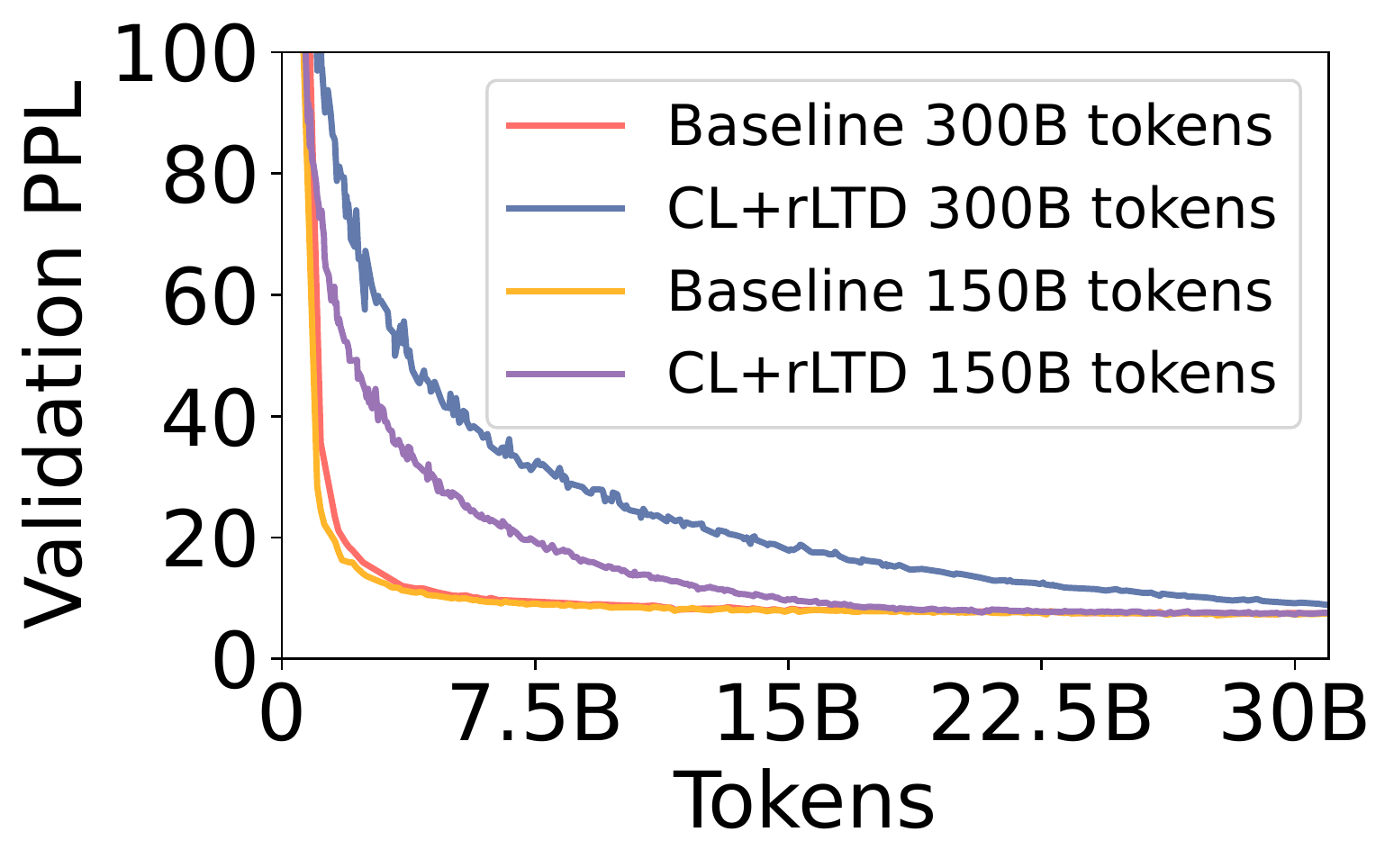}\label{fig_gpt_validation_ppl_1st}}
\subfigure[End of training]{\includegraphics[width=0.46\textwidth]{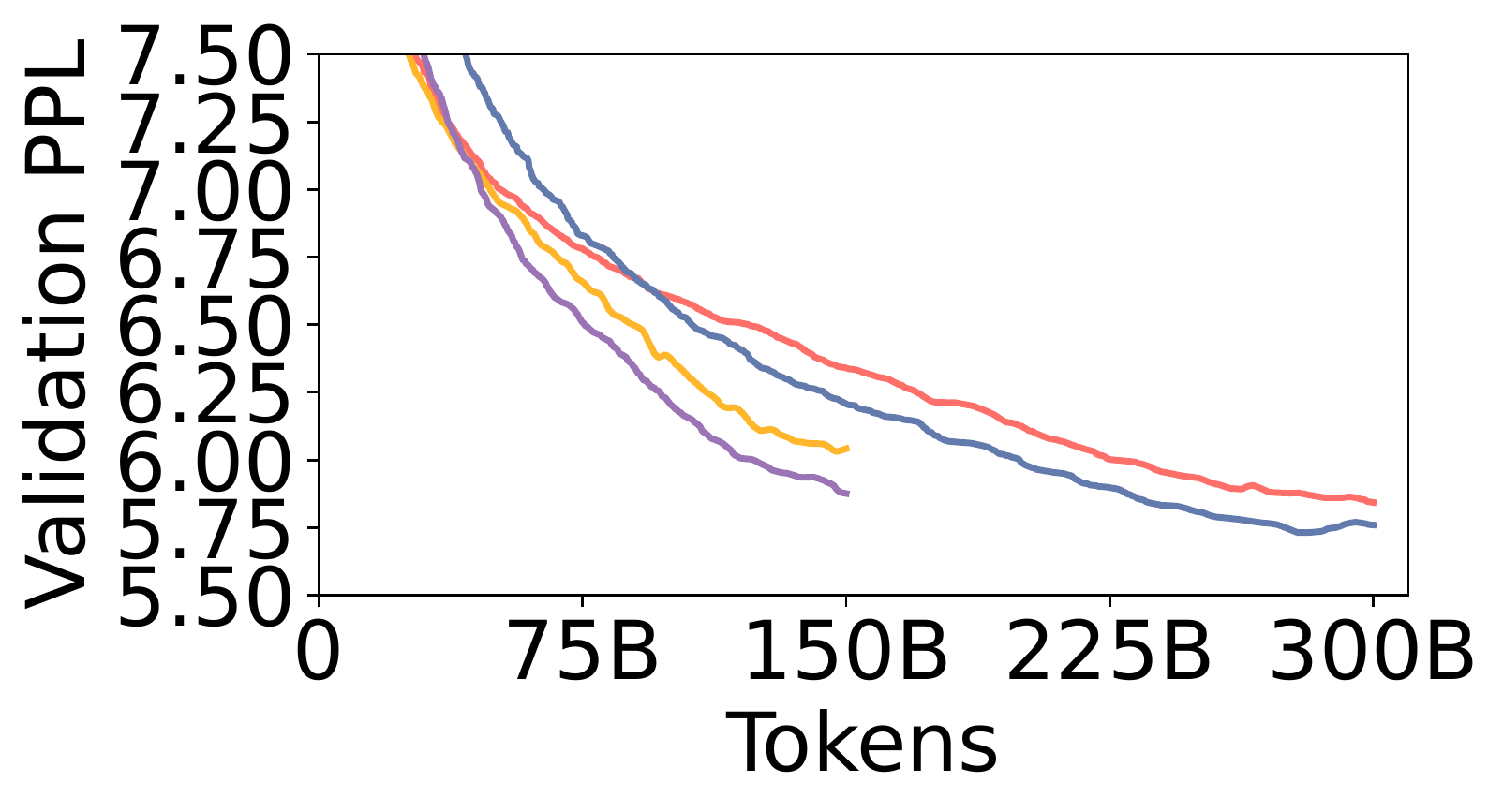}\label{fig_gpt_validation_ppl_2nd}}
\caption{Validation perplexity during GPT-3 1.3B pretraining, comparing the baseline and the best \OURS solution under 100\% and 50\% training data.}
\label{fig_gpt_validation_ppl}
\end{figure*}

\begin{table*}[t]
\setlength{\tabcolsep}{2pt}
\centering
\scriptsize
\begin{tabular}{@{}l|cccccccccccccccc@{}}
\toprule
 && & & & & & & & (8)& & & & & & & (15)\\
 && & & & & (5)& (6)& & CL& & (10)& & & (13)& & CL\\
 && & (2)& (3)& (4)& CL& CL& & seqtru& & CL& & & CL& & seqtru\\
 &(0)& (1)& CL& CL& CL& seqtru& seqres& (7)& +voc& (9)& seqtru& (11)& (12)& seqtru& (14)& +voc\\
Case &OpenAI& baseline& seqtru& seqres& voc& +voc& +voc& rLTD& +rLTD& baseline& +voc& rLTD& baseline& +voc& rLTD& +rLTD\\
Train tokens &300B& 300B& 300B& 300B& 300B& 300B& 300B& 300B& 300B& 200B& 200B& 200B& 150B& 150B& 150B& 150B\\
\midrule
Avg.  & 47.9 & 42.5 & 43.4 & 43.0 & 42.3 & 43.6 & 43.0 & 43.7 & 43.8 & 41.9 & 42.7 & 43.1 & 42.0 & 42.6 & 42.7 & 42.8\\
\midrule
(0) HellaSwag & 54.7 & 51.9 & 52.3 & 52.4 & 51.8 & 52.7 & 52.2 & 54.1 & 54.3 & 50.9 & 52.0 & 52.9 & 49.9 & 50.6 & 51.6 & 52.1\\
(1) LAMBADA & 63.6 & 62.0 & 61.2 & 61.7 & 60.6 & 61.9 & 61.1 & 62.9 & 62.3 & 59.8 & 61.4 & 62.3 & 59.5 & 59.6 & 61.3 & 61.7\\
(2) TriviaQA & 19.7 & 7.0 & 7.91 & 7.63 & 6.66 & 7.65 & 6.07 & 7.9 & 7.55 & 6.15 & 6.46 & 7.54 & 5.9 & 7.2 & 6.37 & 7.42\\
(3) WebQs & 4.63 & 1.38 & 1.62 & 2.07 & 2.56 & 1.38 & 2.02 & 3.15 & 2.17 & 2.46 & 1.67 & 2.31 & 1.03 & 2.26 & 2.66 & 3.2\\
(4) Winogrande & 58.7 & 55.6 & 59.1 & 58.2 & 57.1 & 58.9 & 56.9 & 58.5 & 58.4 & 54.9 & 58.2 & 59.1 & 56.6 & 57.1 & 57.1 & 57.5\\
(5) PIQA & 75.1 & 71.4 & 71.0 & 72.1 & 70.8 & 71.4 & 72.1 & 71.2 & 71.5 & 70.7 & 71.4 & 72.3 & 71.4 & 71.9 & 70.5 & 72.0\\
(6) ARC Challenge & 35.5 & 29.4 & 29.6 & 29.3 & 28.8 & 30.1 & 28.9 & 28.7 & 30.1 & 28.5 & 28.2 & 29.7 & 27.2 & 27.0 & 28.7 & 27.6\\
(7) ARC Easy & 53.8 & 53.7 & 54.3 & 55.0 & 54.0 & 55.2 & 55.0 & 54.4 & 56.4 & 53.5 & 53.2 & 52.7 & 52.7 & 53.7 & 54.1 & 54.0\\
(8) ANLI R1 & 33.4 & 31.6 & 33.3 & 30.7 & 33.4 & 33.5 & 31.6 & 33.0 & 31.6 & 31.6 & 29.8 & 31.9 & 33.0 & 32.9 & 32.1 & 33.7\\
(9) ANLI R2 & 33.3 & 33.7 & 33.8 & 32.8 & 33.0 & 33.3 & 32.9 & 32.5 & 31.5 & 30.4 & 33.2 & 34.8 & 31.8 & 33.9 & 34.6 & 33.6\\
(10) ANLI R3 & 33.4 & 33.1 & 35.2 & 33.5 & 33.2 & 33.3 & 33.9 & 33.4 & 35.2 & 33.7 & 35.8 & 35.3 & 32.4 & 34.8 & 34.9 & 35.0\\
(11) OpenBookQA & 46.8 & 32.4 & 31.8 & 32.0 & 31.2 & 34.0 & 34.6 & 34.0 & 34.0 & 31.0 & 33.0 & 33.8 & 30.4 & 32.4 & 33.6 & 32.4\\
(12) RACE-h & 40.9 & 35.2 & 34.2 & 35.7 & 35.3 & 35.3 & 34.3 & 35.4 & 36.4 & 34.6 & 33.9 & 35.0 & 34.3 & 34.2 & 34.6 & 34.9\\
(13) BoolQ & 62.4 & 62.4 & 63.1 & 62.5 & 60.2 & 62.7 & 63.6 & 61.9 & 63.6 & 62.0 & 62.8 & 61.0 & 61.2 & 59.6 & 61.5 & 61.9\\
(14) Copa & 77.0 & 72.0 & 70.0 & 75.0 & 72.0 & 73.0 & 77.0 & 76.0 & 75.0 & 71.0 & 74.0 & 73.0 & 72.0 & 75.0 & 71.0 & 71.0\\
(15) RTE & 56.0 & 54.2 & 58.1 & 54.9 & 52.0 & 56.0 & 54.2 & 55.0 & 54.5 & 55.2 & 54.9 & 54.2 & 59.2 & 55.6 & 55.2 & 54.5\\
(16) WSC & 61.5 & 36.5 & 42.3 & 36.5 & 34.6 & 43.3 & 36.5 & 43.3 & 40.4 & 36.5 & 37.5 & 36.5 & 36.5 & 36.5 & 37.5 & 36.5\\
(17) MultiRC & 13.6 & 1.05 & 2.1 & 1.47 & 3.15 & 0.944 & 0.944 & 0.839 & 2.41 & 0.839 & 0.839 & 0.839 & 0.839 & 1.68 & 1.05 & 1.15\\
(18) ReCoRD & 85.2 & 83.3 & 83.7 & 83.5 & 83.2 & 83.8 & 83.3 & 84.7 & 84.3 & 82.8 & 82.4 & 84.0 & 82.5 & 82.6 & 83.6 & 83.6\\
\bottomrule
\end{tabular}
\caption{GPT-3 1.3B 0-shot evaluation results. The first column is the results of the original OpenAI GPT-3 1.3B model~\cite{gpt3}. All the other columns are in the same order as the rows in main paper~\tref{tab_gpt_eval}. OpenAI results are not directly comparable to ours because the training data are different.}\label{tab_gpt_0shot_appendix}
\end{table*}

\begin{table*}[t]
\setlength{\tabcolsep}{2pt}
\centering
\scriptsize
\begin{tabular}{@{}l|cccccccccccccccc@{}}
\toprule
 && & & & & & & & (8)& & & & & & & (15)\\
 && & & & & (5)& (6)& & CL& & (10)& & & (13)& & CL\\
 && & (2)& (3)& (4)& CL& CL& & seqtru& & CL& & & CL& & seqtru\\
 &(0)& (1)& CL& CL& CL& seqtru& seqres& (7)& +voc& (9)& seqtru& (11)& (12)& seqtru& (14)& +voc\\
Case &OpenAI& baseline& seqtru& seqres& voc& +voc& +voc& rLTD& +rLTD& baseline& +voc& rLTD& baseline& +voc& rLTD& +rLTD\\
Train tokens &300B& 300B& 300B& 300B& 300B& 300B& 300B& 300B& 300B& 200B& 200B& 200B& 150B& 150B& 150B& 150B\\
\midrule
Avg.  & 49.0 & 44.0 & 44.8 & 44.5 & 44.5 & 44.9 & 44.4 & 44.9 & 45.1 & 44.0 & 44.5 & 44.8 & 42.7 & 43.7 & 43.5 & 44.0\\
\midrule
(0) HellaSwag & 54.9 & 52.4 & 52.7 & 52.6 & 52.0 & 52.7 & 52.8 & 54.7 & 55.1 & 51.2 & 52.2 & 53.4 & 50.5 & 50.9 & 52.2 & 53.0\\
(1) LAMBADA & 57.0 & 57.6 & 56.0 & 57.0 & 55.7 & 57.0 & 57.6 & 59.5 & 59.6 & 55.1 & 56.4 & 58.4 & 54.2 & 55.7 & 57.5 & 58.9\\
(2) TriviaQA & 32.1 & 13.5 & 14.0 & 13.9 & 13.2 & 14.7 & 13.0 & 13.5 & 13.7 & 12.6 & 12.9 & 12.4 & 11.5 & 12.0 & 11.5 & 12.3\\
(3) WebQs & 19.6 & 11.8 & 11.9 & 12.0 & 12.9 & 12.6 & 12.5 & 12.5 & 13.8 & 12.1 & 11.5 & 12.0 & 10.0 & 11.6 & 10.2 & 12.1\\
(4) Winogrande & 59.1 & 57.4 & 56.7 & 58.9 & 58.2 & 60.0 & 58.2 & 58.7 & 58.1 & 55.9 & 59.2 & 59.0 & 56.8 & 58.0 & 58.4 & 58.4\\
(5) PIQA & 74.3 & 71.5 & 71.4 & 71.5 & 71.4 & 71.5 & 72.3 & 71.6 & 72.6 & 71.1 & 72.0 & 71.9 & 71.2 & 71.7 & 71.4 & 71.4\\
(6) ARC Challenge & 36.7 & 32.8 & 32.2 & 33.4 & 32.7 & 32.8 & 32.5 & 32.8 & 34.6 & 32.3 & 32.7 & 33.4 & 31.7 & 31.2 & 30.5 & 31.7\\
(7) ARC Easy & 59.1 & 63.5 & 65.2 & 64.6 & 64.7 & 64.7 & 64.4 & 64.2 & 65.9 & 63.2 & 63.9 & 62.5 & 61.5 & 63.0 & 61.7 & 63.0\\
(8) ANLI R1 & 32.5 & 29.8 & 31.6 & 31.4 & 31.7 & 31.6 & 32.7 & 32.3 & 32.7 & 31.3 & 32.5 & 30.7 & 32.0 & 30.8 & 33.0 & 32.4\\
(9) ANLI R2 & 31.4 & 34.4 & 34.6 & 33.0 & 31.2 & 33.7 & 31.9 & 32.4 & 32.6 & 34.0 & 32.9 & 31.9 & 31.0 & 32.0 & 34.0 & 34.0\\
(10) ANLI R3 & 36.0 & 33.6 & 34.1 & 33.1 & 33.4 & 33.8 & 33.8 & 32.8 & 33.8 & 31.9 & 33.9 & 33.9 & 32.7 & 31.7 & 35.2 & 35.2\\
(11) OpenBookQA & 50.6 & 32.4 & 34.0 & 34.6 & 34.0 & 35.4 & 35.2 & 33.6 & 32.6 & 33.0 & 33.2 & 33.2 & 33.4 & 33.4 & 32.2 & 29.8\\
(12) RACE-h & 41.4 & 34.5 & 36.6 & 35.4 & 35.3 & 36.7 & 35.5 & 37.1 & 36.7 & 35.7 & 34.4 & 35.3 & 35.5 & 34.2 & 35.9 & 34.6\\
(13) BoolQ & 64.1 & 60.8 & 63.5 & 59.4 & 63.1 & 62.1 & 63.1 & 64.2 & 64.0 & 62.8 & 62.1 & 63.8 & 58.8 & 63.4 & 58.2 & 62.0\\
(14) Copa & 77.0 & 76.0 & 74.0 & 79.0 & 76.0 & 76.0 & 74.0 & 73.0 & 74.0 & 74.0 & 77.0 & 76.0 & 69.0 & 70.0 & 71.0 & 70.0\\
(15) RTE & 50.9 & 48.0 & 55.2 & 50.5 & 53.8 & 52.7 & 49.1 & 53.1 & 52.0 & 56.0 & 54.5 & 55.6 & 48.0 & 56.0 & 48.4 & 51.2\\
(16) WSC & 49.0 & 36.5 & 36.5 & 36.5 & 36.5 & 36.5 & 36.5 & 36.5 & 36.5 & 36.5 & 36.5 & 36.5 & 36.5 & 36.5 & 36.5 & 36.5\\
(17) MultiRC & 20.8 & 5.88 & 7.24 & 5.35 & 6.93 & 5.77 & 5.98 & 6.19 & 5.35 & 4.51 & 5.67 & 6.72 & 4.51 & 6.19 & 5.67 & 6.4\\
(18) ReCoRD & 84.0 & 83.0 & 83.4 & 83.3 & 82.4 & 83.6 & 83.2 & 84.6 & 84.0 & 82.3 & 82.7 & 83.9 & 82.2 & 82.4 & 83.8 & 83.3\\
\bottomrule
\end{tabular}
\caption{GPT-3 1.3B 10-shot evaluation results. The first column is the results of the original OpenAI GPT-3 1.3B model~\cite{gpt3}. All the other columns are in the same order as the rows in main paper~\tref{tab_gpt_eval}. OpenAI results are not directly comparable to ours because the training data are different. Note that OpenAI used different number of shots for each task, while we use the same 10 shots for all tasks.}\label{tab_gpt_10shot_appendix}
\end{table*}

\begin{table*}[t]
\setlength{\tabcolsep}{2pt}
\centering
\scriptsize
\begin{tabular}{@{}l|cccccccccccc@{}}
\toprule
 && (2)& & (4)& & (6)& & (8)& & (10)& & (12)\\
 && CL& & CL& & CL& & CL& & CL& & CL\\
 && seqtru& & seqtru& & seqtru& & seqtru& & seqtru& & seqtru\\
 &(1)& +voc& (3)& +voc& (5)& +voc& (7)& +voc& (9)& +voc& (11)& +voc\\
Case &baseline& +rLTD& baseline& +rLTD& baseline& +rLTD& baseline& +rLTD& baseline& +rLTD& baseline& +rLTD\\
Model size &1.3B& 1.3B& 1.3B& 1.3B& 1.3B& 1.3B& 1.3B& 1.3B& 1.3B& 1.3B& 1.3B& 1.3B\\
Train tokens &3B& 3B& 6B& 6B& 12B& 12B& 24B& 24B& 48B& 48B& 96B& 96B\\
\midrule
Avg.  & 34.5 & 35.0 & 36.3 & 36.8 & 37.2 & 38.4 & 38.8 & 40.2 & 39.8 & 41.2 & 41.5 & 42.2\\
\midrule
(0) HellaSwag & 28.7 & 29.3 & 30.8 & 33.2 & 35.4 & 38.1 & 39.0 & 42.7 & 43.5 & 46.9 & 47.8 & 49.9\\
(1) LAMBADA & 28.9 & 32.0 & 38.0 & 41.4 & 43.5 & 49.5 & 50.3 & 53.9 & 54.3 & 58.0 & 57.8 & 60.4\\
(2) TriviaQA & 1.18 & 1.4 & 1.58 & 1.56 & 1.79 & 1.89 & 2.28 & 3.91 & 3.5 & 4.82 & 6.29 & 6.16\\
(3) WebQs & 0 & 0.148 & 0.443 & 0.738 & 1.03 & 0.935 & 0.984 & 0.984 & 1.08 & 2.36 & 2.21 & 2.51\\
(4) Winogrande & 51.3 & 50.8 & 52.2 & 51.0 & 49.5 & 51.8 & 50.7 & 54.1 & 53.5 & 54.9 & 53.3 & 56.5\\
(5) PIQA & 62.1 & 61.6 & 62.5 & 63.5 & 64.9 & 66.6 & 66.8 & 68.5 & 68.6 & 69.6 & 70.1 & 71.3\\
(6) ARC Challenge & 22.2 & 22.9 & 24.9 & 23.0 & 24.7 & 24.6 & 24.1 & 26.2 & 26.7 & 26.6 & 28.5 & 28.2\\
(7) ARC Easy & 38.8 & 38.4 & 40.5 & 41.0 & 44.1 & 45.2 & 46.4 & 47.7 & 48.6 & 50.7 & 51.2 & 52.7\\
(8) ANLI R1 & 33.3 & 33.3 & 32.6 & 33.3 & 31.5 & 31.5 & 31.7 & 32.7 & 33.2 & 33.7 & 33.4 & 33.0\\
(9) ANLI R2 & 33.2 & 34.6 & 35.8 & 32.7 & 31.7 & 32.8 & 32.6 & 33.6 & 33.1 & 34.0 & 34.1 & 34.4\\
(10) ANLI R3 & 32.8 & 33.9 & 35.4 & 32.9 & 34.4 & 34.9 & 35.4 & 34.5 & 32.2 & 35.1 & 33.7 & 33.5\\
(11) OpenBookQA & 25.6 & 24.4 & 26.2 & 27.2 & 28.2 & 28.0 & 28.8 & 29.6 & 30.4 & 31.6 & 32.2 & 31.6\\
(12) RACE-h & 27.1 & 28.5 & 28.9 & 29.4 & 30.0 & 31.2 & 32.2 & 32.5 & 31.8 & 33.5 & 34.5 & 35.2\\
(13) BoolQ & 58.4 & 56.4 & 53.3 & 56.8 & 56.0 & 57.3 & 59.2 & 62.0 & 58.7 & 60.3 & 61.9 & 60.1\\
(14) Copa & 61.0 & 64.0 & 66.0 & 71.0 & 68.0 & 69.0 & 70.0 & 72.0 & 69.0 & 69.0 & 70.0 & 71.0\\
(15) RTE & 52.7 & 52.3 & 53.4 & 53.1 & 53.4 & 54.2 & 54.2 & 53.4 & 52.3 & 53.1 & 53.4 & 55.6\\
(16) WSC & 36.5 & 36.5 & 39.4 & 36.5 & 36.5 & 36.5 & 36.5 & 36.5 & 36.5 & 36.5 & 36.5 & 36.5\\
(17) MultiRC & 0.839 & 0.839 & 1.15 & 0.839 & 1.47 & 0.839 & 0.839 & 0.839 & 0.839 & 1.36 & 0.839 & 1.47\\
(18) ReCoRD & 60.6 & 63.4 & 66.6 & 70.3 & 71.5 & 75.6 & 75.8 & 78.8 & 78.7 & 81.3 & 81.4 & 82.3\\
\bottomrule
\end{tabular}
\caption{GPT-3 1.3B 0-shot evaluation results when pretraining with 1\%, 2\%, 4\%, 8\%, 16\%, and 32\% of data.}\label{tab_gpt_0shot_appendix_2}
\end{table*}

\begin{table*}[t]
\setlength{\tabcolsep}{2pt}
\centering
\scriptsize
\begin{tabular}{@{}l|cccccccccccc@{}}
\toprule
 && (2)& & (4)& & (6)& & (8)& & (10)& & (12)\\
 && CL& & CL& & CL& & CL& & CL& & CL\\
 && seqtru& & seqtru& & seqtru& & seqtru& & seqtru& & seqtru\\
 &(1)& +voc& (3)& +voc& (5)& +voc& (7)& +voc& (9)& +voc& (11)& +voc\\
Case &baseline& +rLTD& baseline& +rLTD& baseline& +rLTD& baseline& +rLTD& baseline& +rLTD& baseline& +rLTD\\
Model size &1.3B& 1.3B& 1.3B& 1.3B& 1.3B& 1.3B& 1.3B& 1.3B& 1.3B& 1.3B& 1.3B& 1.3B\\
Train tokens &3B& 3B& 6B& 6B& 12B& 12B& 24B& 24B& 48B& 48B& 96B& 96B\\
\midrule
Avg.  & 33.9 & 35.0 & 35.6 & 36.6 & 37.3 & 38.8 & 38.8 & 40.7 & 40.7 & 42.3 & 43.0 & 43.2\\
\midrule
(0) HellaSwag & 28.9 & 29.5 & 31.3 & 33.2 & 35.2 & 38.2 & 39.3 & 43.1 & 43.6 & 47.0 & 47.9 & 50.3\\
(1) LAMBADA & 24.5 & 27.5 & 32.2 & 36.2 & 37.6 & 44.9 & 44.0 & 50.7 & 47.0 & 53.2 & 51.8 & 57.0\\
(2) TriviaQA & 0.804 & 1.36 & 1.75 & 3.05 & 3.21 & 4.93 & 5.27 & 6.96 & 7.51 & 9.45 & 10.6 & 11.0\\
(3) WebQs & 1.08 & 1.72 & 2.17 & 2.9 & 3.44 & 5.22 & 4.87 & 6.94 & 7.73 & 8.66 & 10.4 & 11.4\\
(4) Winogrande & 51.6 & 51.0 & 52.2 & 50.2 & 51.8 & 54.0 & 51.7 & 55.2 & 57.0 & 55.1 & 57.0 & 56.1\\
(5) PIQA & 60.9 & 62.0 & 62.1 & 63.9 & 65.3 & 66.5 & 66.0 & 67.9 & 68.8 & 69.7 & 69.8 & 71.1\\
(6) ARC Challenge & 21.9 & 23.2 & 24.0 & 24.3 & 24.8 & 24.9 & 26.5 & 27.7 & 28.0 & 29.8 & 31.5 & 32.1\\
(7) ARC Easy & 38.7 & 41.9 & 44.9 & 47.1 & 50.0 & 52.4 & 54.1 & 55.6 & 56.4 & 59.8 & 60.6 & 62.5\\
(8) ANLI R1 & 31.7 & 33.5 & 33.4 & 32.8 & 34.1 & 32.6 & 35.2 & 33.0 & 31.6 & 33.7 & 33.0 & 31.2\\
(9) ANLI R2 & 33.1 & 35.0 & 30.3 & 34.7 & 35.6 & 34.4 & 34.2 & 31.0 & 33.6 & 34.4 & 32.4 & 32.5\\
(10) ANLI R3 & 33.9 & 34.8 & 35.1 & 33.2 & 33.5 & 34.2 & 33.4 & 33.2 & 34.5 & 33.2 & 34.2 & 32.8\\
(11) OpenBookQA & 25.0 & 26.0 & 27.2 & 28.4 & 28.8 & 26.0 & 27.2 & 28.6 & 29.2 & 31.2 & 32.6 & 33.0\\
(12) RACE-h & 26.9 & 27.8 & 29.1 & 28.9 & 29.1 & 30.5 & 32.3 & 31.9 & 32.0 & 34.3 & 34.4 & 35.0\\
(13) BoolQ & 49.1 & 50.0 & 45.6 & 49.1 & 45.4 & 56.2 & 48.0 & 56.3 & 55.6 & 60.2 & 62.1 & 58.3\\
(14) Copa & 62.0 & 66.0 & 70.0 & 66.0 & 69.0 & 67.0 & 71.0 & 70.0 & 66.0 & 70.0 & 72.0 & 72.0\\
(15) RTE & 53.1 & 49.5 & 47.3 & 50.2 & 48.4 & 48.7 & 48.0 & 56.3 & 55.6 & 50.9 & 54.2 & 49.1\\
(16) WSC & 36.5 & 36.5 & 36.5 & 36.5 & 36.5 & 36.5 & 36.5 & 36.5 & 36.5 & 36.5 & 36.5 & 36.5\\
(17) MultiRC & 5.25 & 4.72 & 4.41 & 4.3 & 5.35 & 4.3 & 5.14 & 3.88 & 5.04 & 5.56 & 5.67 & 6.93\\
(18) ReCoRD & 59.8 & 63.0 & 66.0 & 69.6 & 70.8 & 75.0 & 74.6 & 78.7 & 77.8 & 80.9 & 80.7 & 82.1\\
\bottomrule
\end{tabular}
\caption{GPT-3 1.3B 10-shot evaluation results when pretraining with 1\%, 2\%, 4\%, 8\%, 16\%, and 32\% of data.}\label{tab_gpt_10shot_appendix_2}
\end{table*}

\begin{table}[t]
\centering
\scriptsize
\begin{tabular}{@{}l|cc@{}}
\toprule
 && (2)\\
 && CL\\
 && seqtru\\
 &(1)& +voc\\
Case &baseline& +rLTD\\
Model size &6.7B& 6.7B\\
Train tokens &300B& 300B\\
\midrule
Avg.  & 42.8 & 43.5\\
\midrule
(0) HellaSwag & 53.0 & 53.3\\
(1) LAMBADA & 60.1 & 59.6\\
(2) TriviaQA & 11.0 & 9.31\\
(3) WebQs & 2.95 & 2.31\\
(4) Winogrande & 56.0 & 56.8\\
(5) PIQA & 72.0 & 71.8\\
(6) ARC Challenge & 28.9 & 28.9\\
(7) ARC Easy & 54.5 & 54.2\\
(8) ANLI R1 & 33.6 & 30.8\\
(9) ANLI R2 & 32.8 & 34.1\\
(10) ANLI R3 & 33.6 & 35.5\\
(11) OpenBookQA & 33.6 & 32.4\\
(12) RACE-h & 33.8 & 35.0\\
(13) BoolQ & 61.5 & 57.5\\
(14) Copa & 71.0 & 74.0\\
(15) RTE & 54.5 & 55.2\\
(16) WSC & 36.5 & 51.0\\
(17) MultiRC & 1.89 & 1.78\\
(18) ReCoRD & 82.4 & 82.6\\
\bottomrule
\end{tabular}
\caption{GPT-3 MoE 6.7B 0-shot evaluation results.}\label{tab_gpt_0shot_appendix_3}
\end{table}

\begin{table}[t]
\centering
\scriptsize

\begin{tabular}{@{}l|c@{}}
\toprule
 & Validation loss \\
\midrule
baseline & 8.22 \\
random-LTD (37.76\% token saving)   &   8.26    \\
\tokenbypass (w/ MSLG, 37.76\% token saving) & 9.62  \\
\bottomrule
\end{tabular}
 \caption{Comparing random-LTD and \tokenbypass (both with our proposed MSLG applied) on GPT-3 pretraining.}  \label{tab_tokenbypass_pretrain}
\end{table}

\subsection{GPT-3 pretraining experimental setup and detailed results}
\label{sec:appendix_gpt}
For GPT-3 pretraining, we set some of the hyperparameters the same as the original OpenAI work~\cite{gpt3}: seqlen 2K, batch size 512, learning rate 2e-4 (batch size 256 and learning rate 3e-4 for the GPT-3 MoE 6.7B model since we use 350M as the base model). We set other hyperparameters differently: (1) OpenAI pretrains GPT-3 on 300B tokens. To evaluate data efficiency techniques, we pretrain with 9 different total training tokens: 300B, 200B (67\%), 150B (50\%), 96B (32\%), 48B (16\%), 24B (8\%), 12B (4\%), 6B (2\%), 3B (1\%). (2) When using less than 300B training tokens, we increase the peak learning rate proportionally (e.g., 2x LR when using 50\% data). This is similar to the traditional learning rate scaling when using different batch sizes. However, when using extremely small amount of data (e.g., 1\% data), we find that using too larger learning rate (e.g., 100x) could lead to divergence. In such case we keep halving learning rate until the training succeed. (3) We do not use OpenAI's batch size warmup method because our GPT-3 125M model pretraining experiments show that it does not help on model quality under the same total training tokens. And the smaller batch sizes prevent us to pretrain on large number of GPUs at the beginning, which leads to longer training wall-clock time; (4) Since we don't use the batch size warmup, our training has more tokens at early steps. Thus we increase the linear learning rate warmup duration from OpenAI's 375M tokens to 3B tokens (except when using 3B tokens in total, where we use first 1.5B tokens for warmup); (5) OpenAI uses a single cycle cosine learning rate decay over 260B tokens, and the min learning rate is 10\% of peak learning rate. However, based on our experiments and related works~\cite{yang2022tensor, hoffmann2022training}, we changed the decay duration to always equal to total training token and the min learning rate to always equal to 1e-6, which provide better model quality. When calculating the total consumed training token, we take CL and random-LTD (which change number of tokens on certain steps) into consideration. For CL and random-LTD hyperparameters, we use the low-cost tuning strategy described in~\sref{sec:design}.

To evaluate the quality of pretrained GPT-3 models, we perform 0-shot and 10-shot evaluations on 19 tasks used by original OpenAI work: HellaSwag~\cite{zellers2019hellaswag}, LAMBADA~\cite{paperno2016lambada}, TriviaQA~\cite{joshi2017triviaqa}, WebQs~\cite{berant2013semantic}, Winogrande~\cite{sakaguchi2020winogrande}, PIQA~\cite{bisk2020piqa}, ARC Challenge/Easy~\cite{clark2018think}, ANLI R1/R2/R3~\cite{nie2019adversarial}, OpenBookQA~\cite{mihaylov2018can}, RACE-h~\cite{lai2017race}, BoolQ~\cite{clark2019boolq}, Copa~\cite{afshar2018copa}, RTE~\cite{dagan2013recognizing}, WSC~\cite{levesque2012winograd}, MultiRC~\cite{yadav2019quick}, and ReCoRD~\cite{zhang2018record}. Since there is no additional training involved in 0/10-shot evaluations, it's impossible to try multiple seeds thus each task only has one accuracy result. We then take the average accuracy over the 19 tasks.

Among the 5 CL difficulty metrics we have for GPT-3 model, to find out which metric provides the best model quality we pretrain the model (with 100\% data) 5 times (each with 1 CL metric). 
For seqtru metric (to our knowledge the only metric previously applied to GPT-3 pretraining), we tune the CL hyperparameters $d_s$ and $T_{c}$ based on the tuning strategy proposed in previous work~\cite{slw}. 
Then for other metrics we use the same hyperparameters without retuning for fair comparison. 
As presented in~\tref{tab_gpt_eval} case 1 to 6, results show that all 5 CL metrics provide better model quality than baseline (except (4)CL\_voc's 0-shot accuracy), and the (5)CL\_seqtru\_voc provides the best quality. 
The extensibility of our general CL library enables us to easily apply different CL metrics to this large-scale model pretraining with huge training data, and identify a new CL metric that provides better model quality than existing solution (2)CL\_seqtru. 
Next we pretrain the model with 67\% data, comparing the baseline and the best CL metric we find. 
Results show that the average 0-shot evaluation accuracy drops from 42.5 to 41.9 when baseline use less data (\tref{tab_gpt_eval} case 1, 9). 
On the other hand, our CL solution (case 10) with 67\% data is able to achieve better 0-shot and 10-shot accuracy than baseline with 100\% data, achieving a 1.5x data and time saving.

When applying the proposed random-LTD technique, results show similar benefit as CL: better model quality when using 100\% data (\tref{tab_gpt_eval} case 7), and 1.5x data/time saving while maintaining model quality (case 11). 
To explore whether composing CL and random-LTD could achieve even better data and training efficiency, first we pretrain the model with both techniques under 100\% training data. 
Results (case 5, 7, 8) show that using both techniques together further improves the model quality, demonstrating the benefit of composability by our framework. 
Next we pretrain the model with 50\% data. Results (case 12 to 15) show that the baseline has worse 0-shot and 10-shot evaluation accuracy under 2x less data. 
Using CL or random-LTD can only recover part of the accuracy loss. 
On the other hand, the composed data efficiency solution is able to achieve the same or better accuracy results as baseline with 100\% data, demonstrating a 2x data and 2x time saving.

To better understand how the proposed approach influences the model convergence,~\fref{fig_gpt_validation_ppl} plots the token-wise validation perplexity during pretraining. 
At the beginning of the training the proposed approach has slower convergence since we focus on easier/simpler data samples (CL) and drop more tokens (random-LTD) at the beginning. 
On the other hand, at the later stage of training the proposed approach is able to provide faster convergence speed than baseline. 
Our approach with 50\% data is able to achieve similar final validation perplexity as baseline with 100\% data (while baseline with 50\% data cannot). 
Our approach with 100\% data is able to achieve even better final validation perplexity which leads to the highest model quality.

The proposed methods' slower convergence in~\fref{fig_gpt_validation_ppl_1st} does not mean that "the proposed methods have limited performance compared to the baseline when operating at a smaller data scale". This is because no matter how the data scale changes, the best configurations (the number of CL/random-LTD steps) of the proposed methods also change in proportion (as summarized in~\tref{table_guide}). Thus no matter how small the data scale/total data budget is, the proposed methods will only have slower convergence at the early stage of that training, yet still provide better final model quality/training efficiency after the full training. The end result will always be similar as~\fref{fig_gpt_validation_ppl_2nd}, regardless of data scale. As shown in~\fref{fig:moti_pareto}, we did test and demonstrate that the proposed methods provide better final model quality/training efficiency at a wide range of data scales from 3B to 300B. In terms of which of the techniques contribute more to the convergence slowdown at the early stage of training, our results show that CL contributes more to the slowdown. This makes sense because compared to CL which completely focus on easier data, random-LTD still have first and last layer acting as normal layers without token dropping.


As presented in~\sref{sec:intro} and~\fref{fig:moti_pareto}, we also compare baseline and proposed work when using even less data during GPT-3 pretraining (Detailed accuracy results can be found in Appendix~\ref{sec:appendix_gpt}). Results show that our approach provides better model quality at all cost budgets, advancing the whole cost-quality Pareto frontier. In particular, we achieve up to 12.5x data/time/cost saving while still maintaining 95\% of the model quality (zero-shot eval accuracy) compared to the baseline with full data. 
Based on measured training time, this would be a cost reduction from \$46.3K to \$3.7K if renting similar hardware on Azure~\cite{azure}, greatly democratizing research and usage of foundation models.

Recent work shows that applying Mixture-of-Experts (MoE) to GPT-style model pretraining could lead to better training efficiency while reaching similar model quality~\cite{rajbhandari2022deepspeed}. Thus we also pretrain a GPT-3 MoE 6.7B model (350M base model, together with 64 experts on every other layer) to compare baseline and proposed work. Results show that MoE model does achieve similar model quality with less training cost (\tref{tab_gpt_eval} case 1, 16). On the other hand, our approach can further improve MoE model's model quality (case 17), confirming its broad applicability. 

\tref{tab_gpt_0shot_appendix} and~\ref{tab_gpt_10shot_appendix} present the 0-shot and 10-shot accuracy results for each task achieved by the pretrained GPT-3 1.3B models. \tref{tab_gpt_0shot_appendix_2} and~\ref{tab_gpt_10shot_appendix_2} present the 0-shot and 10-shot accuracy results for the same GPT-3 1.3B model but pretrained with even less data as discussed in main paper \fref{fig:moti_pareto}, \sref{sec:intro}, and \sref{sec:eval-gpt}. \tref{tab_gpt_0shot_appendix_3} presents the 0-shot accuracy results for each task achieved by the pretrained GPT-3 MoE 6.7B models, as discussed in main paper \sref{sec:eval-gpt}.

\begin{table*}[t]
\setlength{\tabcolsep}{1pt}
\centering
\scriptsize
\begin{tabular}{@{}l|ccccccccccc@{}}
\toprule
Case & Train tokens & Average & MNLI-m & MNLI-mm & QQP & QNLI & SST-2 & CoLA & STS-B & MRPC & RTE \\
\midrule
(0)original & 43B &82.1 &86.7 & 85.9 &72.1 &92.7 &94.9 &60.5 &86.5 &89.3 &70.1  \\
(1)baseline &1049B &87.29±0.53 &88.54±0.16 & 89.25±0.13 &92.1±0.07 &94.12±0.15 &94.33±0.48 &64.36±1.59 &90.43±0.21 &89.32±0.81 &83.2±1.16  \\
(2)CL\_seqtru &1049B &87.31±0.57 &89.03±0.14 & 89.35±0.24 &92.21±0.03 &94.12±0.11 &94.68±0.1 &62.08±2.06 &90.72±0.27 &89.58±0.52 &83.98±1.64  \\
(3)CL\_seqreo &1049B &87.48±0.61 &88.81±0.16 & 89.27±0.19 &92.2±0.12 &93.99±0.28 &94.79±0.42 &62.86±1.85 &90.51±0.25 &89.32±0.85 &85.55±1.34  \\
(4)CL\_voc &1049B &87.36±0.64 &88.64±0.23 & 89.24±0.16 &92.32±0.05 &94.03±0.09 &95.14±0.31 &63.34±1.82 &90.07±0.18 &89.84±1.06 &83.59±1.83  \\
(5)CL\_seqtru\_voc &1049B &87.6±0.34 &88.9±0.1 & 89.29±0.17 &92.26±0.05 &94.26±0.19 &95.25±0.4 &64.6±0.6 &90.38±0.25 &90.62±0.22 &82.81±1.05  \\
(6)CL\_seqreo\_voc &1049B &87.06±0.52 &88.73±0.13 & 88.91±0.26 &92.32±0.07 &93.92±0.08 &94.91±0.25 &61.05±1.15 &90.36±0.23 &89.32±1.13 &83.98±1.34  \\
(7)random-LTD &1049B &88.17±0.48 &88.74±0.25 & 89.18±0.21 &92.27±0.1 &94.32±0.21 &95.02±0.38 &67.3±1.5 &90.65±0.15 &90.1±0.63 &85.94±0.89  \\
(8)CL\_seqtru\_voc+random-LTD &1049B &87.69±0.32 &88.79±0.13 & 89.26±0.04 &92.34±0.08 &94.21±0.23 &95.14±0.36 &65.46±0.68 &90.44±0.19 &89.58±0.56 &83.98±0.59  \\
(9)baseline &703B &87.19±0.49 &88.75±0.18 & 89.11±0.19 &92.13±0.08 &93.99±0.16 &95.14±0.46 &62.07±1.44 &90.08±0.31 &89.84±0.68 &83.59±0.87  \\
(10)CL\_seqtru\_voc &703B &87.29±0.62 &88.96±0.07 & 89.15±0.25 &92.21±0.09 &94.23±0.08 &95.25±0.33 &62.19±1.75 &89.92±0.21 &90.1±0.55 &83.59±2.25  \\
(11)random-LTD &703B &87.99±0.38 &88.86±0.1 & 88.79±0.12 &92.01±0.12 &94.25±0.17 &94.68±0.32 &67.1±0.9 &90.55±0.19 &89.32±0.39 &86.33±1.12  \\
(12)baseline &524B &86.61±0.5 &88.53±0.14 & 88.77±0.17 &92.04±0.11 &93.93±0.19 &95.02±0.25 &61.05±1.22 &89.88±0.25 &88.28±1.08 &82.03±1.13  \\
(13)CL\_seqtru\_voc &524B &86.9±0.33 &88.66±0.14 & 89.25±0.21 &92.08±0.05 &93.99±0.26 &95.02±0.17 &63.34±0.52 &89.96±0.25 &88.54±0.22 &81.25±1.14  \\
(14)random-LTD &524B &87.32±0.48 &88.81±0.15 & 88.9±0.13 &91.96±0.04 &94.28±0.14 &94.91±0.43 &64.41±1.32 &90.39±0.25 &89.06±0.18 &83.2±1.67  \\
(15)CL\_seqtru\_voc+random-LTD &524B &87.44±0.46 &88.9±0.19 & 88.9±0.13 &92.19±0.09 &94.17±0.12 &94.68±0.35 &65.97±1.09 &90.31±0.22 &89.06±0.79 &82.81±1.13  \\
\bottomrule
\end{tabular}
\caption{BERT-large finetuning results. The first row is the results of the original BERT-large model~\cite{bert}. All the other rows are in the same order as the rows in main paper~\tref{tab_bert_finetune}. Original BERT results are not directly comparable to ours because the training data and total training token are different.}\label{tab_bert_finetune_appendix}
\end{table*}

\subsection{BERT-large pretraining experimental setup and results}
\label{sec:appendix_bert}
For BERT-large pretraining, we set some of the hyperparameters the same as the Megatron-LM work~\cite{megatron} since it achieves better model quality than original BERT: seqlen 512, batch size 1024, learning rate 1e-4 with linear warmup up at first 10000 steps and then linearly decay to 1e-5. We set other hyperparameters differently: (1) Megatron-LM pretrains over 2M steps (1049B tokens).
To evaluate data efficiency techniques, we pretrain with 3 different total training tokens: 1049B, 703B (67\%), and 524B (50\%). 
(2) When using less than 1049B training tokens, we increase the peak learning rate proportionally. (3) Megatron-LM decays the learning rate over 2M steps. Since our techniques could change the number of tokens at some steps, we change the decay to token-based and set the decay duration always the same as total training tokens. For CL and random-LTD hyperparameters, we use the low-cost tuning strategy described in~\sref{sec:design}.

To evaluate the quality of pretrained BERT-large models, we finetune the models for 8 tasks from the GLUE benchmark~\cite{glue}: MNLI, QQP, QNLI, SST-2, CoLA, STS-B, MRPC, RTE. We follow the finetuning hyperparameters from the original BERT work~\cite{bert}: 3 epochs, batch size 32. For learning rate we test {5e-5, 4e-5, 3e-5, 2e-5} on the baseline and find that 3e-5 provides the best average GLUE score, thus we select LR=3e-5 for the comparison between baseline and proposed work. We perform finetuning on 5 seeds (1234 to 1238) and take the median/std on each task, then we take the average of the median scores as the average GLUE score, and take the average of std scores as the overall std.

Similar to the GPT-3 case, for CL we first investigate which metric (among 5 metrics we have for BERT model) provides the best model quality by pretraining the model (with 100\% data) 5 times. 
\tref{tab_bert_finetune} case 1 to 6 results show that 4 CL metrics provide better model quality than baseline, and the (5)CL\_seqtru\_voc provides the best quality. 
Next we pretrain with 67\% data, comparing the baseline and our best CL metric. 
Results show that the GLUE score drops from 87.29 to 87.19 when baseline use less data (case 1, 9). 
On the other hand, our CL solution (case 10) with 67\% data is able to achieve on-par GLUE score as baseline with 100\% data, achieving a 1.5x data and time saving.

\tref{tab_bert_finetune} case 7, 11, 14 present the case when applying random-LTD only. 
In terms of data saving random-LTD performs better than CL: it is able to achieve better GLUE score even with 2x less data than baseline (case 14), greatly surpassing the 1.33x data saving by the state-of-the-art \tokenbypass method. 
However, the time saving is less than data saving because the token dropping mechanism adds a computation overhead at each step. 
Because the BERT-large is a smaller model than GPT-3 1.3B, this fixed latency overhead has a larger relative impact to the training time. 
However, even with this overhead random-LTD is still a more data/time-efficient solution than baseline/\tokenbypass.

\tref{tab_bert_finetune} case 8 and 15 present the case when applying both CL and random-LTD. 
At 50\% data, the composed solution further improves the GLUE score from the CL/random-LTD-only cases (case 15), achieving a 2x data and 1.8x time saving while maintaining the GLUE score compared to baseline. 
Another thing to note is that this case also has more time saving than the random-LTD-only case. 
This is because CL will first truncate the sequences before random-LTD perform the random token selection, and the shorter sequences reduces random-LTD's computation overhead. 
At 100\% data, the composed solution (case 8) improves the GLUE score from the CL-only case, but is worse than the random-LTD-only case. 
One hypothesis is that for BERT pretraining when composing the two techniques it's preferable to reduce the CL duration, but exhaustively testing all hyperparameters is out of our resource budget and this work's scope.

\tref{tab_bert_finetune_appendix} presents the finetuning results for each task achieved by the pretrained BERT-large models.

\subsection{GPT-2 finetuning experimental setup}
\label{sec:appendix_ptb}

Due to the lack of published training recipe, we first perform a hyperparameter search for the baseline case (256 combinations of batch size, LR schedule, number of epochs). 
Then using the combination that provides best baseline validation perplexity, we apply CL and random-LTD (each with 16 different combinations of their two hyperparameters) to verify if they could further improve the model quality. 

For \gpthf finetuning on PTB~\cite{marcus-etal-1993-building}, we use an already-pretrained \gpthf model checkpoint and an example script~\footnote{\url{https://github.com/huggingface/transformers/blob/main/examples/pytorch/language-modeling/run_clm_no_trainer.py}} from Huggingface. 
Given the much smaller training cost (about 38min on a single V100 for 5 epochs), we focus on improving the model quality under the same amount of data. 
Due to the lack of published training recipe, we first perform a hyperparameter search for the baseline case: we tried 256 combinations of batch size (4, 8, 16, 32), learning rate (2e-5, 3e-5, 5e-5, 10e-5), learning rate warmup (0\% and 10\% linear warmup steps), learning rate decay (no decay, linear decay), and number of epochs (2, 3, 5, 10). 
For this sweep we only use one seed (1234) due to the number of combinations. 
Results show that the best combination among the 256 cases is: batch size 4, learning rate 10e-5, 0\% learning rate warmup, linear learning rate decay, and 5 epochs. 
Results also show that for this task using more epochs (5 or 10) leads to better validation perplexity than less epochs (2 or 3). 

Then using this combination that provides best baseline validation perplexity, we apply CL and random-LTD (each with 16 different combinations of their two hyperparameters) to verify if they could further improve the model quality. 
For CL we test 5 metrics (seqtru, seqres, voc, seqtru\_voc, seqres\_voc), each with 16 different combinations of its two hyperparameters: start difficulty $d_s$ (8, 32, 128, 512 for seqtru/seqres, and 1\%, 10\%, 30\%, 50\% for voc) and total CL steps $T_{c}$ (10\%, 30\%, 50\%, 70\% of the baseline's total steps). 
Results show that the seqres metric provides the best model quality, and its best hyperparameter combination is $d_s = 32, T_{c} = 70\%$ of baseline steps. 
For random-LTD we test 16 different combinations of its two hyperparameters: start seqlen $r_s$ (8, 32, 128, 512) and total steps $T_{r}$ (10\%, 30\%, 50\%, 70\% of the baseline's total steps). 
Results show that the best hyperparameter combination is $r_s = 128, T_{r} = 30\%$ of baseline steps. 
For CL+random-LTD composed case, we re-tuned the combination of $T_{c}$ and $T_{r}$ (CL will first adjust seqlen before random-LTD. 
To have a meaningful composition, it essentially requires $T_{c} < T_{r}$) and the best combination is $d_s = 32, r_s = 128, T_{c} = 10\%, T_{r} = 30\%$ of baseline steps. 
At last, for the best case of baseline, CL, random-LTD, and CL+random-LTD, we run another 4 seeds (1235 to 1238) and then calculate the median/std of the validation perplexity.

As shown in \tref{tab_gpt_finetune}, seqres provides the best model quality among the 5 CL metrics (case 3), unlike the two pretraining tasks where the seqtru\_voc is the best metric. 
This is because this finetuning task has much smaller batch size and number of tokens per batch. 
seqtru will reduce number of tokens per batch, which is less desirable under small-batch training. 
The small batch also prevents the voc metric to include sufficient number of samples with different vocabulary rarity, limiting its benefit. 
Applying random-LTD also improves the model quality (case 7). 
Both CL best metric and random-LTD are able to surpass baseline on all 16 combinations of their hyperparameters, demonstrating that they are not sensitive to the hyperparameter choices. 
At last we try another 4 seeds for the baseline, CL best metric, random-LTD, and the CL+random-LTD case. 
The composed CL+random-LTD case (case 8) further improves model quality from random-LTD-only case, but is only on-par with CL-only case. 
One hypothesis is that for tasks with such small-scale training data, it's less possible to further improve model quality by composing multiple data efficiency techniques.

\subsection{ViT finetuning experimental setup}
\label{sec:appendix_vit}
We apply random-LTD to the vision transformer (\vit)~\citep{dosovitskiy2021an} on finetuning tasks to demonstrate the broader applications of our method across different domains. 
We use the pretrained models published in~\citep{rw2019timm} and test on two small image recognition benchmarks--- CIFAR10 and CIFAR100~\citep{krizhevsky2009learning}, and one large-scale dataset---ImageNet~\citep{deng2009imagenet}. 
For ImageNet (CIFAR10/100), we use the 12-layer (24-layer) pretrained \vit with an input resolution $224\times 224$ in which each patch of size $16\times 16$ such that the sequence length becomes $196+1$ (the extra token is for position). 
ImageNet (CIFAR10/100) is trained on 8-GPU (1-GPU) and the batch size is 32 (128) images per GPU. 
The training budget for all three datasets is 14 epochs and a small constant learning rate is used based on grid search.
Particularly, the best learning rate for ImageNet (CIFAR) is 5e-5 (1e-4). 
For ImageNet (CIFAR), when applying random-LTD the sequence length is started with 66 (32) and linearly reaches to the 197 full sequence length at 80\% of the total baseline training iterations, equivalent to a 1.3x (1.4x) data saving. Detailed experiment results are summarized in~\tref{table:vit-main-full}.

\begin{table}[t]
\centering
\scriptsize
\begin{tabular}{@{}l@{\hskip 0.01in}|@{\hskip 0.01in}c@{\hskip 0.01in}|@{\hskip 0.01in}c@{\hskip 0.01in}|@{\hskip 0.01in}c@{}}
\toprule
& \multicolumn{3}{c}{CIFAR datasets on 24-layer ViT}            \\
& Data saving & Top-1 (CIFAR100) & Top-1 (CIFAR10) \\
\midrule
baseline   & N/A          & 93.93±0.30    & 99.32±0.05     \\
random-LTD & 1.4x     & 94.02±0.40    & 99.30±0.03 \\
\midrule
& \multicolumn{3}{c}{ImageNet datasets on 12-layer ViT}            \\
& Data saving & Top-1       & Top-5\\
\midrule
baseline   & N/A            & 84.65±0.04 & 97.41±0.02      \\
random-LTD & 1.3x     & 84.70±0.04  & 97.48±0.02  \\
\bottomrule
\end{tabular}
 \caption{ViT finetuning results.} \label{table:vit-main-full}
\end{table}

\subsection{Comparing random-LTD with the \tokenbypass work}
\label{sec:appendix-tokenbypass}
In main paper~\sref{sec:eval-bert} we demonstrate that random-LTD achieves 2x data saving while maintaining model quality for BERT pretraining, greatly surpassing the 1.3x data saving achieved by the state-of-the-art \tokenbypass work~\cite{hou-etal-2022-token}. 
In this section we provide additional discussion and evaluation to compare random-LTD with \tokenbypass.

\begin{figure*}[t]
    \centering
     \includegraphics[width=0.8\linewidth]{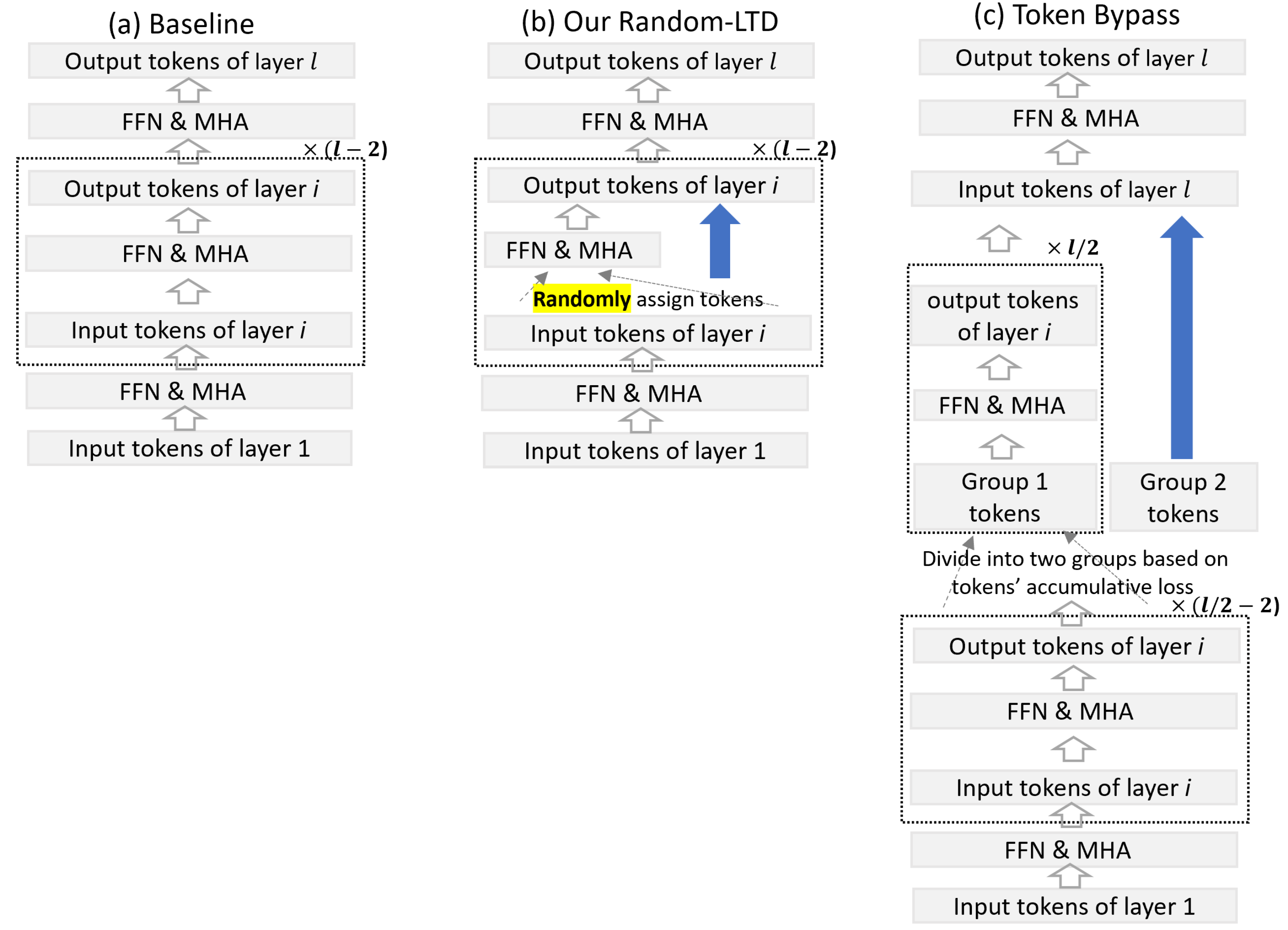}
    \caption{Illustration of the transformer model for the baseline training (left),  \tokenbypass training (right) and random-LTD training (middle). Compared to \tokenbypass, random-LTD requires no criterion on the dropped tokens and trains well for all middle layers. The box with dash line is a repeated block. For both (a) and (b), the block is repeated by $l-2$ times, while for (c), the block is repeated by $l/2$. In the box, ``Output tokens of layer $i$'' is the same as ``Input tokens of layer $i+1$''.   }\label{fig:illustration_of_random_ltd_and_baseline_and_tokenbypass}
\end{figure*}

We include the illustration of the comparison between baseline, \tokenbypass, and random-LTD in~\fref{fig:illustration_of_random_ltd_and_baseline_and_tokenbypass}. 
First, the takeaway from \tokenbypass can be summarized into (1) drop unimportant tokens starting from an intermediate layer of the model, (2) the dropping schedules is a fixed constant function (drop half of the tokens), and (3) the dropping criterion based on the ``accumulated masked language modeling loss'' (which is referred to as ``token loss'' since it needs each token's loss)

However, \tokenbypass have several limitations (1) only tested on BERT pretraining (we find that it’s less effective in GPT pretraining and finetuning), (2) the bypass layer starting only from an intermediate layer (e.g., 6L for BERT-base), and (3) the dropping criterion based on “token loss” may not be accessible for some tasks, like classification problems.

Acknowledging that we are inspired by their excellent work and trying to solve their limitations, we believe random-LTD consists of three differences: (1) drop tokens starting from the 2nd layer of the model, (2) propose a linear increasing dropping schedule to close the training and inference discrepancy, and (3) the new random dropping criterion (which has lower overhead and can be easily applied to tasks without “token loss”, such as vision transformer). Next, we provide more direct comparisons between random-LTD and \tokenbypass on GPT-2 finetuning and GPT-3 pretraining tasks. Note that because this study was performed in parallel with other experiments, the hyperparameter choices are different from the experiments in main paper.

\textbf{GPT-2 finetuning on PTB with various constant dropping schedule.} 
To better demonstrate the benefit of random selection per layer, we provide a study with various constant dropping schedule. Particularly, from the second layer to the last second layer, we use one of the sequence lengths from {921, 819, 716, 614, 512, 409}, of which the corresponding token saving ratio are shown in~\tref{tab_tokenbypass_fixed}. We finetune \gpthf (24 layers) on the PTB dataset with constant learning rate 5e-5 and Adam optimizer for 15 epochs (batch-size 8). The results are the best validations (average of three runs and one standard deviation) of random-LTD (without Monotonic Sequence Length Growth, MSLG) and \tokenbypass.

\begin{table*}[t]
\centering
\scriptsize
\begin{tabular}{@{}l|cccccc@{}}
\toprule
Token saving ratio & 1.88\% & 12.75\% & 23.72\% & 34.59\% & 45.45\% & 56.43\% \\
\midrule
random-LTD (w/o MSLG)   & 16.15±0.01&	16.83±0.06&	17.95±0.08&	20.02±0.05&	23.35±0.16&	30.65±0.78      \\
\tokenbypass & 16.4±0.04&	17.3±0.06&	18.59±0.19&	23.09±0.23&	28.56±0.24&	35.91±0.26  \\
\bottomrule
\end{tabular}
 \caption{Comparing random-LTD (w/o MSLG) and \tokenbypass under various constant dropping schedule. Baseline achieves a perplexity of 16.11±0.04.} 
 \label{tab_tokenbypass_fixed}
\end{table*}

As shown in~\tref{tab_tokenbypass_fixed}, for all cases random-LTD has better performance than \tokenbypass, even without one of the key contributions, Monotonic Sequence Length Growth (MSLG). 
This further verifies the conjecture we made in the main paper: ``However, several works~\citep{vig2019analyzing,michel2019sixteen,voita2019analyzing} have shown that \mha focuses on different tokens at different layer depths and the attention map aligns with the dependency relation most strongly in the middle of transformer architectures. 
Therefore, \tokenbypass used in~\cite{hou-etal-2022-token}, i.e.,  fully skipping middle layers, may hinder the learnability/generalization of the architecture during pretraining/inference.''

\textbf{GPT-2 finetuning on PTB with our proposed MSLG.} 
We are also curious if MSLG can help boost the performance of \tokenbypass. 
Therefore, we also perform the comparison between random-LTD (with MSLG) and TokenBypass (with MSLG) on GPT-2 finetuning. 
We start at sequence length from 128 and linearly increase to full sequence 1024, with a different total steps to achieve different token saving ratios shown in~\tref{tab_tokenbypass_mslg}. 
The rest of the hyperparameters are the same as the previous experiment. 

\begin{table*}[t]
\centering
\scriptsize

\begin{tabular}{@{}l|cccccccc@{}}
\toprule
Token saving ratio & 8\% & 16\% &24\% &32\% &40\% &47\% &52\% &55\% \\
\midrule
random-LTD   & 15.91±0&15.86±0.06&15.86±0.01&15.85±0.02&16.05±0.06&17.02±0.05&18.41±0.04&20.01±0.06      \\
\tokenbypass (w/ MSLG) &16.1±0.02&16.09±0.05&16.21±0.03&16.54±0.01&17.06±0.04&18.64±0.04&23.12±0.22&25.77±0.57  \\
\bottomrule
\end{tabular}
 \caption{Comparing random-LTD and \tokenbypass (both with our proposed MSLG applied) under various token saving ratios. Baseline achieves a perplexity of 16.11±0.04.} 
  \label{tab_tokenbypass_mslg}
\end{table*}

Note that under MSLG it is hard to control the overall token saving ratio to be the same as constant dropping schedule case. 
But comparing~\tref{tab_tokenbypass_mslg}’s 24\%/47\%/55\% with~\tref{tab_tokenbypass_fixed}’s 23.72\%/45.45\%/56.43\%, we can clearly see the benefit of MSLG.
Meanwhile, comparing the results of random-LTD and TokenBypass (with MSLG), it is clear that random-LTD still has better performance than TokenBypass for all cases. 
This shows that the other components of random-LTD, particularly the layerwise dropping mechanism, has its unique advantage over accumulated token loss for auto-regressive generative models.

\textbf{GPT-3 pretraining.} 
To directly compare the two techniques on pretraining tasks, we pretrain a GPT-3 350M model with 30B tokens. 
Due to limited time and resource, this is a smaller model and 10\% of data compared to our other GPT-3 pretraining experiments. 
And due to the same reason we only compare the validation loss at the end of pretraining, but our experience shows that this metric has strong correlation with downstream task zero/few-shot evaluation performance. 
Based on the last GPT-2 finetuning experiment, here we again apply MSLG to \tokenbypass. 
Results in~\tref{tab_tokenbypass_pretrain} shows that under the same token saving ratio, random-LTD provides significantly better model quality than \tokenbypass.

\textbf{Other downstream tasks.} 
\tokenbypass cannot be easily extended to various downstream tasks. 
The reason is that the \tokenbypass criterion is based on the ``token loss'', but downstream tasks, e.g., classification and regression (GLUE benchmark), do not have ``token loss''. 
Therefore, we did not find an easy way to apply \tokenbypass on those tasks.

\subsection{Additional related work discussions}
\label{sec:appendix-related}

We believe our work is orthogonal to the findings in the Scaling Law work~\cite{scaling-law}. In the Scaling Law work, the are 4 key findings related to data: (1) Model performance depends strongly on the size of the dataset. (2) Model size and data size have to be increased simultaneously in order to consistently achieve better model quality. (3) Large models are more sample-efficient than small models, reaching the same level of performance with fewer optimization steps. (4) Under a fixed computation budget, large model with less data can lead to better performance. In our work, \fref{fig:moti_pareto} reconfirms their finding 1, and our overall experience agree with their other 3 findings. Our proposed methods aim to further improve the data efficiency, but the overall relationship between data and model performance still holds the same.

Drop path~\cite{huang2016deep} is a technique commonly used in neural architecture search where random paths of layers are dropped during training. We believe the Drop path work is more like a special case of our work. In Drop path, the whole mini batch are skipped for a subset of layers every time. In our work, we only skip a subset of tokens for each training sample at each layer. We believe Drop path's complete dropping a subset of layers could lead to worse convergence and/or less stable training for the modern Transformer-based models, as we discussed in~\sref{sec:design-ltd}. Furthermore, the \tokenbypass work mentioned in our paper also make some tokens fully skip a subset of layers (similar to Drop path), and in Appendix~\ref{sec:appendix-tokenbypass} we provided a thorough comparison between random-LTD and the existing work \tokenbypass. Results show that random-LTD provides better benefits on both GPT-2 finetuning and GPT-3 pretraining tasks.

EViT~\cite{liang2022not} and Peeling the Onion~\cite{kong2023peeling} are two additional related data routing techniques. We believe that our work provide sufficient contributions beyond these two related works: First, both works only verify their methods on ViT models, while our methods are verified on both NLP and CV large-scale models. Second, both works provide less benefit than our work. When zero model quality degradation is required, the EViT work can only provide 1.15x training speed for ImageNet training (Table 11 in their paper). In contrast, in our work~\tref{table:vit-main-full}, our random-LTD method can provide 1.3x training speedup while slightly improving the model quality. The Peeling the Onion work achieves 1.15x training speedup while slightly improving the model quality, which is again less speedup gain than our random-LTD. Furthermore, when combining both of our proposed methods we demonstrate even better training speedup gain.

\end{document}